\documentclass{svjour3}
\usepackage[hyphens]{url}
\usepackage{xspace}
\usepackage{color}
\usepackage{array}
\usepackage{underscore}
\usepackage{tipa}
\usepackage{cite}
\usepackage{stmaryrd}

\usepackage{rotating}

\usepackage{amsmath}
\usepackage{amssymb}
\usepackage{booktabs}
\usepackage{graphicx}
\usepackage{listings}
\usepackage{paralist}
\usepackage{subfig}
\usepackage{hyperref}
\usepackage{breakurl}
\usepackage{footnote}

\makeatletter

\def\holdocspecials{\do\ \do\$\do\&%
  \do\#\do\^\do\^^K\do\_\do\^^A\do\%}

\def\holtt{\trivlist \item[]\if@minipage\else\vskip\parskip\fi
\leftskip\@totalleftmargin\rightskip\z@
\parindent\z@\parfillskip\@flushglue\parskip\z@
\@tempswafalse \def\par{\if@tempswa\hbox{}\fi\@tempswatrue\@@par}
\obeylines \tt \let\do\@makeother \holdocspecials
 \frenchspacing\@vobeyspaces}

\makeatother

\newlength{\hsbw}
\newcommand\HOLSpacing{13pt}

\newenvironment{holnb}{\begin{flushleft}
 \begin{minipage}[b]{\hsbw}
 \vspace*{.06in}
 \begingroup\small\baselineskip\HOLSpacing\footnotesize
 \begin{holtt}}{\end{holtt}\endgroup
 \end{minipage}
 \end{flushleft}}

   \newcommand\hilbert{\varepsilon}
   
   \newcommand{\Thm}{\(\vdash\)}
   \newcommand{\Cond}{\(\rightarrow\)}
   \newcommand{\Eqv}{\(\equiv\)}
   \newcommand{\Iff}{\(\Longleftrightarrow\)\hspace{-1.5mm}}
   \newcommand{\Fa}{\(\forall\)}
   \newcommand{\Et}{\(\exists\)}
   \newcommand{\Eu}{\(\exists_{unique}\)}

   \newcommand{\Impl}{\(\Longrightarrow\)\hspace{-1.5mm}}
   \newcommand{\Func}{\(\to\)\hspace{-1.5mm}}

   \newcommand{\Lam}{\(\lambda\)}
   
   \newcommand{\Minus}{\(-\)}
   \newcommand{\Lminus}{\(-\)\hspace{-1.5mm}}
   \newcommand{\Prime}{\('\)}
   \newcommand{\Und}{\_}
   \newcommand{\Lt}{\(<\)}
   \newcommand{\Gt}{\(>\)}
   \newcommand{\Leq}{\(\leq\)}
   \newcommand{\Geq}{\(\geq\)}
   \newcommand{\Eq}{\(=\)}
   \newcommand{\Lrb}{\((\)}
   \newcommand{\Rrb}{\()\)}

   \newcommand{\Next}{\(\bigcirc\)}
   \newcommand{\Prev}{\(\ominus\)}
   \newcommand{\WPrev}{\(\widetilde{\bigcirc}\)}
   \newcommand{\Event}{\(\Diamond\)}
   \newcommand{\Once}{\(\underline{\Diamond}\)}  
\newcommand{\Hilbert}{\(\hilbert\)}

\newcommand{\Conj}{\(\wedge\)}
\newcommand{\Disj}{\(\vee\)}
\newcommand{\Neg}{\(\neg\)}
\newcommand{\Pnd}{\(\Diamond\)}

\newcommand{\Empt}{\(\varnothing\)}
\newcommand{\Models}{\(\models\)}

\long\def\holthm#1{{\def\Turns{\Thm} \rechol#1\end\end\end}}

\long\def\rechol#1#2#3{\let\next=\rechol\def\postnext{#2#3}\ifx#1\end
\let\next=\relax\def\postnext{\relax}
\else\ifx#1!\Fa                                          
\else\ifx#1@\Hilbert                                     
\else\ifx#1\#\Pnd                                        
\else\ifx#1'\Prime                                       
\else\ifx#1~\Neg                                         
\else\ifx#1\~\Neg
\else\ifx#1_\Und                                         
\else\ifx#1(\ifx#2+\ifx#3)\Next\def\postnext{}\fi        
            \else\ifx#2-\Prev\def\postnext{}             
            \else\ifx#2~\ifx#3)\WPrev\def\postnext{}\fi            
             \else\Lrb\fi\fi\fi                          
\else\ifx#1)\Rrb%
\else\ifx#1\/\Disj                                       
\else\ifx#1\.\Lam                                        
\else\ifx#1>\ifx#2=\Geq\def\postnext{#3}\else\Gt\fi      
\else\ifx#1?\ifx#2!\Eu\def\postnext{#3}\else\Et\fi       
\else\ifx#1-\ifx#2>\Func\def\postnext{#3}               
            \else\ifx#2-\Lminus\def\postnext{#3}
            \else\Minus\fi\fi                               
\else\ifx#1|\ifx#2-\Turns\def\postnext{#3}               
            \else\ifx#2=\Models\def\postnext{#3}
                 \else\Bar\fi\fi
\else\ifx#1<\ifx#2=\ifx#3>\Iff\def\postnext{}       
                   \else\Leq\def\postnext{#3}\fi    
            \else\ifx#2+\Event\def\postnext{}       
            \else\ifx#2-\Once\def\postnext{}       
            \else\Lt\fi\fi\fi                       
\else\ifx#1=\ifx#2=\ifx#3>\Impl\def\postnext{}            
                   \else\Eqv\def\postnext{#3}\fi         
            \else\ifx#2>\Cond\def\postnext{#3}
                 \else\Eq\fi\fi
\else\ifx#1/\ifx#2\^^M\Conj\par\def\postnext{#3}         
            \else\ifx#2\ \Conj\ \def\postnext{#3}\else#1\fi\fi  
\else#1\fi\fi\fi\fi\fi\fi\fi\fi\fi\fi\fi\fi\fi\fi\fi\fi\fi\fi\fi
\expandafter\next\postnext}

\newcolumntype{*}{>{\global\let\currentrowstyle\relax}}
\newcolumntype{^}{>{\currentrowstyle}}
\newcommand{\rowstyle}[1]{\gdef\currentrowstyle{#1}%
  #1\ignorespaces
}

\def\systemname#1{\textsf{#1}\xspace}

\def\th#1{\texttt{#1}\xspace}

\newcommand{\HOLLight}{\systemname{HOL Light}}
\newcommand{\HOL}{\systemname{HOL}}
\newcommand{\ProverNine}{\systemname{Prover9}}
\newcommand{\Ivy}{\systemname{Ivy}}
\newcommand{\Otter}{\systemname{Otter}}
\newcommand{\Isabelle}{\systemname{Isabelle}}
\newcommand{\Sledgehammer}{\systemname{Sledgehammer}}
\newcommand{\MetiTarski}{\systemname{MetiTarski}}
\newcommand{\Mizar}{\systemname{Mizar}}
\newcommand{\MPTP}{\systemname{MPTP}}
\newcommand{\Vampire}{\systemname{Vampire}}
\newcommand{\Epar}{\systemname{Epar}}
\newcommand{\Z}{\systemname{Z3}}
\newcommand{\E}{\systemname{E}}
\newcommand{\V}{\systemname{V}}
\newcommand{\SUMO}{\systemname{SUMO}}
\newcommand{\Cyc}{\systemname{Cyc}}
\newcommand{\Coq}{\systemname{Coq}}
\newcommand{\MML}{\systemname{MML}}
\newcommand{\MizAR}{\systemname{Miz$\mathbb{AR}$}}
\newcommand{\SPASS}{\systemname{SPASS}}
\newcommand{\Flyspeck}{\systemname{Flyspeck}}
\newcommand{\SNoW}{\systemname{SNoW}}

\newcommand{\MoMM}{\systemname{MoMM}}
\newcommand{\ILF}{\systemname{ILF}}
\newcommand{\PSE}{\systemname{PS-E}}
\newcommand{\SInE}{\systemname{SInE}}
\newcommand{\MaLARea}{\systemname{MaLARea}}
\newcommand{\MaLeCoP}{\systemname{MaLeCoP}}
\newcommand{\Why}{\systemname{Why3}}
\newcommand{\Yices}{\systemname{Yices}}
\newcommand{\CVC}{\systemname{CVC3}}
\newcommand{\AltErgo}{\systemname{AltErgo}}
\newcommand{\OCaml}{\systemname{OCaml}}
\newcommand{\LEO}{\systemname{LEO2}}
\newcommand{\Satallax}{\systemname{Satallax}}
\newcommand{\iProver}{\systemname{iProver}}
\newcommand{\Paradox}{\systemname{Paradox}}
\newcommand{\leanCoP}{\systemname{leanCoP}}
\newcommand{\Metis}{\systemname{Metis}}

\newcommand{\MESON}{\texttt{MESON}\xspace}

\newenvironment{mypar}[2]
  {\begin{list}{}%
    {\setlength\leftmargin{4mm}
    \setlength\rightmargin{4mm}}
    \item[]}
  {\end{list}}

\title{Learning-Assisted
Automated Reasoning with \Flyspeck}
\author{Cezary Kaliszyk \and Josef Urban}
\institute{Cezary Kaliszyk, supported by FWF grant P26201
\at University of Innsbruck, Austria \and
  Josef Urban, funded by NWO grant \textit{Knowledge-based Automated  Reasoning} 
\at  Radboud University, Nijmegen.}

\begin{document}
\maketitle
\begin{abstract}
The considerable mathematical knowledge encoded by the \Flyspeck
project is combined with external automated theorem provers (ATPs) and
machine-learning premise selection methods trained on the \Flyspeck proofs, 
producing an AI system 
capable of proving a wide range of mathematical
conjectures automatically. The 
performance of this architecture
is evaluated in a bootstrapping scenario emulating the development of
\Flyspeck from axioms to the last theorem, each time using only the
previous theorems and proofs. It is shown that 39\% of the 14185 
theorems could be proved in a push-button mode (without any
high-level advice and user interaction) in 30 seconds of real time on
a fourteen-CPU workstation.

The necessary work involves: (i) an implementation of sound
translations of the \HOLLight logic to 
ATP formalisms: untyped
first-order, polymorphic typed first-order, and typed higher-order,
(ii) export of the dependency information from \HOLLight and ATP proofs for
the machine learners, and (iii) choice of suitable representations and
methods for learning from previous proofs, and their integration as
advisors with \HOLLight.
This work
is described and discussed here, 
and an initial analysis of the
body of proofs that were found fully automatically is provided.

\end{abstract}

\newpage
\section{Introduction and Motivation}
\label{introduction}

\begin{mypar}{10mm}{5mm}
{\it
``It is the view of some of us that many people who could have easily contributed to project QED have been distracted away by the enticing lure of AI or AR.'' 

{\hfill\rm-- The QED Manifesto}
}
\end{mypar}

\begin{mypar}{10mm}{5mm}

{\it

 ``So it will take 140 man-years to create a good basic library for formal mathematics.''

{\hfill\rm-- Freek Wiedijk~\cite{Wiedijk01}} 

}

\end{mypar}

\begin{mypar}{10mm}{5mm}

{\it

``We will encourage you to develop the three great virtues of  a programmer:\\ laziness, impatience, and hubris.''

{\hfill\rm-- Larry Wall, Programming Perl~\cite{Wall2000}} 

}

\end{mypar}

\begin{mypar}{10mm}{5mm}
{\it

``And in demonstration itself logic is not all. The true mathematical
reasoning is a real induction [...]''

{\hfill\rm-- Henri Poincar\'e, Science and Method~\cite{Poincare13}} 

}

\end{mypar}

\subsection{Large-Theory Automated Reasoning and \HOLLight}

Use of external first-order automated theorem provers (ATPs) like \Vampire~\cite{Vampire},
\E~\cite{Sch02-AICOMM}, \SPASS~\cite{WeidenbachDFKSW09}, and recently
also SMT (satisfiability modulo theories) solvers like \Z~\cite{z3} for (large-theory)
formalization has been developed considerably in the recent
decade. Particularly in the \Isabelle community, the
\Sledgehammer~\cite{BlanchetteBN11,BlanchetteBP11} bridge to such external tools is getting
increasingly popular. This helps to further develop various parts of
the technology involved. ATPs have recently gained the ability to quickly load
large theories over large signatures and work with them~\cite{HoderV11}. Methods for
automated selection of relevant knowledge and for proof guidance are
actively developed~\cite{Urban11-ate}, together with specialized automated systems
targeted at particular mathematical domains~\cite{AkbarpourP10,PW06,Beeson01}.  Formats and translation
methods handling more formalization-friendly foundations are being
defined~\cite{SutcliffeSCB12,GelderS06,tff1}, and metasystems that decide which ATP, translation method,
strategy, parallelization, and premises to use to solve a given
problem with limited resources are being designed~\cite{sledgehammer10,US+08}. Cooperation of
humans and computers over large corpora of formal knowledge is an
interesting field, allowing exploration of new AI systems and combinations
of different AI techniques that can attempt to encode 
concepts like
analogy and intuition, and rigorously evaluate their usefulness.
Perhaps not only Hilbert and Turing, but also the
formality-opposing and intuition-oriented Poincar\'e\footnote{2012 is not just the year
of Turing~\cite{Hales12}, but also of Poincar\'e, whose ideas about creativity and
invention involving random, intuition-guided exploration confirmed by critical
evaluation quite correspond to what AI systems like \MaLARea~\cite{US+08}
try to emulate in large formal theories.}~\cite{Poincare13} would have been interested to learn about the
new ``semantic AI paradise'' of such large corpora of fully
computer-understandable mathematics (from which we do not intend to be
expelled).

The \HOLLight~\cite{Harrison96} system is probably the first among the existing
well-known interactive theorem provers (ITPs) which has integrated and extensively used a general
ATP procedure, the \MESON tactic~\cite{Har96}.   Hurd has developed and
benchmarked early bridges~\cite{Hurd99,Hurd02} between \HOL and
external systems, and his \Metis system~\cite{hurd2003d} has also
become a significant part of the \Isabelle/\Sledgehammer bridge to
ATPs~\cite{PaulsonS07}. Using the very detailed \Otter/\Ivy~\cite{MS00} proof
objects, Harrison also later implemented a bridge from \HOLLight to \ProverNine~\cite{McC-Prover9-URL}.
\HOLLight however does not yet have a general bridge to large-theory
ATP/AI (``hammer''\footnote{Larry Paulson is guilty of introducing  this
  ``striking'' terminology.}) methods, similar to \Isabelle/\Sledgehammer or \MizAR~\cite{abs-1109-0616,UrbanS10}, which would
attempt to automatically solve a new goal by selecting relevant
knowledge from the large library and running (possibly 
customized/trained)
external ATPs on such 
premise
selections. \HOLLight seems to be a natural candidate for adopting such
methods, because of the amount of
work already done in this direction mentioned above, and also thanks
to \HOLLight's foundational closeness to \Isabelle/\HOL. Also, 
thanks to the \Flyspeck project~\cite{Hales05}, \HOLLight is becoming less of  a
``single, very knowledgable formalizer'' tool, and is getting increasingly used as a ``tool for
interested mathematicians'' (such as the \Flyspeck team in Hanoi\footnote{\url{http://weyl.math.pitt.edu/hanoi2009/Participants/}}) 
who may know the
large libraries much less and have 
less experience with crafting their own 
proof tactics. For
such ITP users it is good  to provide a small number of strong methods
that allow fast progress, which can perhaps also complement the declarative
modes~\cite{abs-1201-3601} pioneered by \HOLLight~\cite{Harrison96a} in the LCF world.

\subsection{\Flyspeck as an Interesting Corpus for Semantic AI Methods}

The purpose of the \Flyspeck project is to produce a formal proof of
the Kepler Conjecture~\cite{Kepler11,HalesHMNOZ10}.
The \Flyspeck development (which in this paper always means
also the required parts of the \HOLLight library) is an interesting corpus
for a number of reasons. First, it formalizes considerable parts of
standard mathematics, and thus exposes a large body of
interconnected mathematical reasoning to all kinds of semantic AI methods and
experiments. Second, the formalization is done in a relatively
directed way, with the final goal of the Kepler conjecture in mind. For
example, in the \Mizar library\footnote{\url{www.mizar.org}} (and even more in other collections like
the \Coq contribs\footnote{\url{http://coq.inria.fr/V8.2pl1/contribs/bycat.html}}), articles may be contributed as isolated
developments, and only much later (or never) re-factored into a form
that makes them work well with related developments. Such refactoring
is often a nontrivial process~\cite{RudnickiT03}. In a directed
development like \Flyspeck, such integrity is a concern from the very
beginning, and this concern should result in the theorems working
better together to justify new conjectures that combine the areas covered by the
development. Third, the language of \HOLLight is in a certain sense simpler than the
language of \Mizar and \Coq (and to a lesser extent also than
\Isabelle/\HOL), where one typically first needs to set up the right
syntactic/type-automation environment to be able to formulate new
conjectures in advanced areas. This greater simplicity (which may come at a cost)
 makes it
possible to write direct (yet advanced) queries to the AI/ATP 
(``hammer'') 
system in the original language, without much additional need for specifying the
context of the query. This could make such ``hammer'' more easy to try for
interested mathematicians, and allow them to explore formal mathematics and \Flyspeck. 
And fourth, \Flyspeck is accompanied with
an informal (\LaTeX) text that is often cross-linked to the formal
concepts and theorems developed in \HOLLight. With sufficiently strong automated reasoning
over the library, this cross-linking opens the way to experiments with
alignment (and eventual semi-automated translation) between the informal
and formal \Flyspeck texts, using corpus-driven methods for language
translation, assisted by such an AI/ATP ``hammer'' as an additional
semantic filter/advisor.

\subsection{The Rest of the Paper}
The work reported here makes several steps towards the above goals:
\begin{enumerate}
\item Sound and efficient
translations of the \HOLLight formulas to several ATP (TPTP)
formalisms are implemented (Section~\ref{translation}). This includes the untyped first-order 
(FOF) format~\cite{SutcliffeSCG06}, the polymorphic typed first-order (TFF1) format~\cite{tff1}, and the typed
higher-order (THF) format~\cite{GelderS06,SB10}.
\item Dependency information is exported from the \Flyspeck
  proofs (Section~\ref{export}). This allows experiments with
  re-proving of theorems by 17 different ATPs/SMTs from their \HOLLight dependencies, and
provides an initial dataset for machine learning of premise selection from previous proofs.
\item Several feature representations characterizing \HOLLight formulas are
  proposed, implemented, and used for machine-learning of premise selection. Several
  preprocessing methods are developed for the dependency data that are
  used for learning. The trained premise-selection systems are
  integrated as external advisors for \HOLLight. A prototype 
  system answering real-time mathematical queries by running various
  parallel combinations of the premise selectors and ATPs is built and
  made available as an online service. See Section~\ref{premiseselection}.
\end{enumerate}
The methods are evaluated in Section~\ref{Experiments}, and it is
shown that by running in parallel the most complementary proof-producing methods on a
14-CPU workstation, one now has a 39\% chance to prove the next
\Flyspeck theorem within 30 seconds in a fully automated push-button
mode (without any high-level advice). 50\% of the \Flyspeck
theorems can be re-proved within 30 seconds by a collection of 7 ATP
methods (run in parallel) if the \HOLLight proof dependencies are
used. 56\% of the theorems could be proved by the union of all methods
tried in the evaluation. 
An initial analysis of these sets of proofs
is given in Section~\ref{comparison}. It is shown that the proofs
produced by the learning-advised ATPs can occasionally develop ideas
that are very different from the original \HOLLight proofs, and that
the learning-advised ATPs can sometimes produce simpler proofs and discover duplications
in the library. Section~\ref{related} discusses related work and
Section~\ref{future} suggests future directions.

\section{Translation of \HOLLight Formulas to ATP formats}
\label{translation}

The HOL logic differs from the formalisms used by most of the existing
ATP and SMT systems. The main differences to first-order logic are the
use of the polymorphic type system, and higher-order features
(guarded by the type system) such as quantification (abstraction) over
higher-order objects and currying. On the other hand, the logic is
made classical and comes with a straightforward intended interpretation
in ZFC. Translation of this logic (and its type-class extension used
by \Isabelle/\HOL) to ATP formalisms has been an active research topic
started already in the 90s. Prominent techniques, such as lambda lifting, 
suitable type system translation methods, etc., have been
described several times ~\cite{Hurd99,Hurd02,Har96,MengP08,BlanchettePhd}.
Therefore this section assumes familiarity with
these techniques, and only briefly
summarizes the logic and the translation approaches considered, and
their particular suitability for the experiments over the \HOLLight
corpora.  For a comprehensive recent overview and discussion of this
topic and the issues related to the translation see Blanchette's
thesis~\cite{BlanchettePhd}. In particular, it contains the arguments about the soundness and (in)completeness of the translation methods that we eventually chose.

\subsection{Summary of the HOL Logic}
\HOLLight uses the \textit{HOL logic}~\cite{Pitts93}: an extended
variant of Church's simple type theory~\cite{Church:SimplyTyped}. Type
variables (implicitly universally quantified) are explicitly added to
the language (providing polymorphism), together with arbitrary type
operators (constructors of compound types like \texttt{`int list'} and
\texttt{`a set'}).  In the HOL logic, the terms and types are intended
to have a standard set-theoretical interpretation in HOL universes. A
\textit{HOL universe} $U$ is a set of non-empty sets, such that $U$ is
closed under non-empty subsets, finite products and powersets, an
infinite set $I \in U$ exists, and a choice function $ch$ over $U$
exists (i.e., $\forall X \in U: ch(X) \in X$) .  The subsets,
products, and powersets together also yield function spaces. A
frequently considered example of a HOL universe is the set
$V_{\omega+\omega} \setminus \{0\}$,\footnote{
$V_{\omega+\omega}$ is the $\omega+\omega$-th set of von Neumann's (cumulative)
hierarchy of sets obtained by iterating the powerset operation
starting with the empty set $\omega+\omega$ times.
This shows that the HOL
  logic is in general weaker than ZFC.} with $ch$ being its
(ZFC-guaranteed) selector, and $I = \omega$.  The standard $U$-interpretation of a
\textit{monomorphic} (i.e., free of type variables) type $\sigma$ is a set
$\llbracket\sigma\rrbracket \in U$,
a \textit{polymorphic} (i.e., containing type variables) type $\sigma$ with $n$ type variables is interpreted as
a function $\llbracket\sigma\rrbracket:U^n \rightarrow U$, and the arrow operator
observes the standard function-space behavior (lifted to appropriate
mappings for polymorphic types) on the type interpretations.  The
standard interpretation of a closed monomorphic term $t:\sigma$ is an element
of the set $\llbracket\sigma\rrbracket \in U$, and a closed polymorphic term (with $n$
type variables) $t:\sigma$ with $\llbracket\sigma\rrbracket:U^n \rightarrow U$ is
interpreted as a (dependently typed) function assigning to each
$n$-tuple $[X_1, \dots , X_n] \in U^n$ an element of $\llbracket\sigma\rrbracket([X_1,
\dots , X_n])$.  The HOL logic's type signature starts with the
built-in nullary type constants \texttt{ind}, interpreted as the infinite set $I$, and
\texttt{bool} (type of propositions), interpreted as a chosen two-element set in $U$ (its
existence follows from the properties of a HOL universe). The term
signature initially contains the polymorphic
constants 
$=_{\alpha \rightarrow \alpha \rightarrow bool}$, and
$\epsilon_{(\alpha \rightarrow bool) \rightarrow \alpha}$, interpreted
as the equality and selector on each set in $U$.  The inference
mechanisms start with a set of standard primitive inference rules,
later adding the axioms of functional extensionality, choice (implying
the excluded middle in the HOL setting), and infinity.  New type and
term constructors can be introduced by simple definitional extension
mechanisms, which are in \HOLLight also used to introduce the standard
logical connectives and quantifiers. The result is a classical logic
system that is in practice quite close to set theory, differing from
it mainly by the built-in type discipline (allowing also complete automation
of abstraction) and by more frequent use of total functions to model
mathematical objects. For example, predicates are modelled as total functions to
\texttt{bool} on types, and sets are in \HOLLight identified with (unary) predicates.
The main issues for translation are the type
system and the automated reification (abstraction) mechanisms that
are not immediately available in first-order logic and may be encoded
in more or less efficient and complete ways. 

\subsection{The \MESON Translation}
\label{meson}

An obvious first idea for generating FOL ATP problems from \HOLLight problems was to re-use parts of the already implemented \MESON tactic.
This tactic tries to justify a given goal $G$ with a supplied list of premises $P_1, ... , P_n$ by
calling a customized first-order ATP implemented in \HOLLight, which is based on the
model elimination method invented by Loveland~\cite{Loveland68}, 
later combined with a Prolog-like search tree~\cite{Loveland78}.  The
implementation of the \MESON tactic in \HOLLight first applies a number of standard translation techniques (such as $\beta$-reduction followed by lambda lifting, skolemization, 
introduction of the \texttt{apply} functor,\footnote{Identity is used by \MESON as the \texttt{apply} functor.} etc.)
that transform the HOL goal (together with the supplied premises) to a clausal FOL goal (or multiple goals). An
interesting (and \MESON-specific) part of the transformation is a rather exhaustive and heuristic instantiation
of the (often polymorphic) premises (called \th{POLY_ASSUME_TAC}), described below.  The clausal FOL
goal is then passed to the core ATP. If the core ATP succeeds, it returns a proof, which is then
translated into \HOLLight proof steps.  
The
transformation from HOL to FOL is heuristic, incomplete, and tuned for
relatively small problems. An interesting feature of \MESON is that the
core ATP does not treat equality specially (as is quite common in tableau
provers), which in turn allows using multiple instantiated versions of
equality (e.g., on lists and on real numbers) inside one problem. Such
equational separation, when combined with the heuristic instantiation
of other polymorphic constants done by \MESON, then prevents the core
ATP from doing ill-typed inferences without the necessity for any
additional type guards. 

The most interesting part of the translation heuristically
instantiates the (possibly polymorphic) premises $P_1, ... , P_n$ and
adds them to the goal $G$. This is done iteratively, building a new
temporary goal $G_i$ (where $G_0=G$) from each premise $P_i$ and the
previous goal $G_{i-1}$ as follows. All (possibly polymorphic)
constants are collected from $P_i$ and $G_{i-1}$, and the set of all
their pairs is created. When such a pair $\{c_P,c_G\}$ consists of two
(symbolically) equal constants, the type of $c_P$ is matched to the
type of $c_G$, and if a substitution $\sigma$ exists (i.e., $Type_{c_P}\sigma =
Type_{c_G}$), it is added to the resulting set $\Sigma_i$ of
type substitutions. Each type substitution from $\Sigma_i$ is then
applied to $P_i$, and all such resulting instances of $P_i$ are added
as assumptions to $G_{i-1}$, yielding $G_i$.
The set of assumptions of the goal $G_{i}$ is thus typically greater
than that of $G_{i-1}$, and the same typically holds for the set of
constants in $G_{i}$, which will be in turn used to instantiate
$P_{i+1}$.

This procedure is quite effective for the small problems that \MESON
normally handles. However, for problems with many premises and many
polymorphic constants this turns out to be  very
inefficient. While re-using \MESON allowed the quick initial exploration
of using external ATPs and advisors described in~\cite{KaliszykU12}, this inefficiency
practically excluded the (seemingly straightforward) use of the
unmodified \MESON procedure as an (at least basic) translation method for generating ATP
problems with many premises. This is why the experiments presented here use 
different translations, described below.

Completeness of the translation from HOL to FOL is in general hard to achieve in an efficient way. 
The \MESON translation is incomplete in several ways. The goal's proper
assumptions are not monomorphised, and the free variables of
polymorphic types are not used in the same way as the polymorphic constants. For example, given the premise:

$$\forall P : \alpha \to \mbox{bool}.\; \forall x : \alpha.\; P x$$

\noindent and a goal that does not mention $\alpha$,
the premise will never be instantiated to the type present in the
goal, and thus will not be usable for \MESON.

\subsection{Translation to the TFF1 and FOF Formats}
\label{tff1}

There is a ``simple'' solution to the instantiation blow-up
experienced with the \MESON translation: avoid heuristic instantiation
as a pre-processing step, and instead let the ATPs handle it as a part
of the ATP problems. This technique is used
in the \Mizar/\MPTP translation~\cite{Urban06,Urb04-MPTP0,Urban03}, where the (dependent and
undecidable) soft type system cannot be separated from the core
predicate logic. The relevant heuristics can instead be developed (and
experimented with) on the level of ATPs. Indeed, for example the
\SPASS system includes a number of
ATP techniques for both complete and incomplete work with
(auto-detected) types~\cite{Weidenbach+99,BlanchettePWW12}. This approach has been in the recent years
facilitated by developing type-aware TPTP standards such as TFF0, TFF1,
and THF, which -- unlike related type-aware efforts like DFG~\cite{Hahnle96} and KIF~\cite{GF92}
-- seem to be more successful in being adopted by ATP and tool
developers. In the case of the recent TFF1 standard~\cite{tff1} adding \HOL-like
polymorphic types to first-order logic, a translation tool to the FOF
and SMT formats has been developed in 2012 by Andrei Paskevich as part
of the \Why system~\cite{FilliatreM07}, simplifying the first experiments with the
non-instantiating translation.

The translation to TFF1 proceeds similarly to the \MESON translation,
but without applying the \th{POLY_ASSUME_TAC}. The problem
formulas are $\beta$-reduced, the remaining lambda abstractions are
again removed using lambda lifting, and the \texttt{apply} functor is
heuristically introduced. The particular heuristic for this is the one
used by Meng and Paulson, i.e., for each higher-order constant $c$ the
minimum arity $n_c$ with which it appears in a problem is computed,
and the first $n_c$ arguments are always passed to $c$ directly inside
the problem. If the constant is also used with more arguments in the
problem, \texttt{apply} is used. Blanchette~\cite{BlanchettePhd} reports that this
optimization works fairly well for \Isabelle/\Sledgehammer, and gives
a simple example when it introduces incompleteness.  As an
example of the translation to the TFF1 format, consider the re-proving
problem\footnote{By a \emph{re-proving problem}, we mean the ATP problem
  consisting of the translated \HOLLight theorem together with the premises used
in its original \HOLLight proof.} for
the theorem \th{Float.REAL_EQ_INV}\footnote{\url{http://mws.cs.ru.nl/~mptp/hh1/OrigDepsProbs/i/p/09895.p}} 
 
proved as
part of the Jordan curve theorem formalization,\footnote{\url{http://mws.cs.ru.nl/~mptp/hol-flyspeck/trunk/Jordan/float.html\#REAL_EQ_INV}} 
whose \HOLLight proof is as follows:

\begin{holnb}\holthm{
let REAL_EQ_INV = prove(`!x y. ((x:real = y) <=> (inv(x) = inv (y)))`,
((REPEAT GEN_TAC))
THEN (EQ_TAC)
THENL [((DISCH_TAC THEN (ASM_REWRITE_TAC[])));
 (* branch 2*) ((DISCH_TAC))
THEN ((ONCE_REWRITE_TAC [(GSYM REAL_INV_INV)]))
THEN ((ASM_REWRITE_TAC[]))]);;
}\end{holnb}
The dependency tracking (see Section~\ref{dependencies}) has found the
following dependencies of the 
theorem:\footnote{\th{Tactics_jordan.unify_exists_tac_example} is just \th{`T=T`}. The
name is accidental.}

\begin{verbatim}
AND_DEF FORALL_DEF IMP_DEF REAL_INV_INV REFL_CLAUSE TRUTH 
Tactics_jordan.unify_exists_tac_example
\end{verbatim}
From these dependencies, only \th{REAL_INV_INV} has nontrivial
first-order content (a list of the trivial facts has been collected and is used for such filtering). 
The problem creation additionally adds three facts encoding
properties of (\HOL) booleans, and also the functional extensionality axiom (\th{EQ_EXT}). In
the original \HOLLight syntax the re-proving problem looks as follows:

\begin{verbatim}
%   ORIGINAL: Float.REAL_EQ_INV
% Assm: EQ_EXT: !f g. (!x. f x = g x) ==> f = g
% Assm: BOOL_CASES_AX: !t. (t <=> T) \/ (t <=> F)
% Assm: NOT_CLAUSES_WEAK_conjunct1: ~F <=> T
% Assm: REAL_INV_INV: !x. inv (inv x) = x
% Assm: TRUTH: T
% Goal: !x y. x = y <=> inv x = inv y
\end{verbatim}
After applying $\beta$-reduction, lambda lifting (none in the example), and introducing the
\texttt{apply} functor (called here \texttt{happ}), this is
transformed (still as \HOL terms) into the following:

\begin{verbatim}
%   PROCESSED
% Assm: !f g. (!x. happ f x = happ g x) ==> f = g
% Assm: !t. (t <=> T) \/ (t <=> F)
% Assm: ~F <=> T
% Assm: !x. inv (inv x) = x
% Assm: T
% Goal: !x y. x = y <=> inv x = inv y
\end{verbatim}
The application functor \texttt{happ} was only used for the function
variables in the extensionality axiom (\th{EQ_EXT}). The function
\texttt{inv} is always used with one argument in the problem, so it is
never wrapped with \texttt{happ}. Finally, the TFF1 TPTP export
declares the signature of the symbols and type operators, and adds the
corresponding guarded quantifications to the formulas. The
\texttt{apply} functor is called \texttt{i} in the TFF1 export (for concise
output in case of many applications in a goal), and it
explicitly takes also the type arguments (A and B in
\th{aEQu_EXT}). This (making the implicit type variables
explicit) is in TFF1 done for any symbol that remains polymorphic. 
We reserve the predicate \texttt{p} for translation between Boolean terms and formulas. 
This is done in the same way as in~\cite{MengP08}. 

\HOLLight allows one identifier to denote several different underlying
constants. In the running example, \texttt{inv} is such an overloaded identifier and denotes
the inverse operations on several different types. To deal with such
identifiers different names are used for each underlying constant
separately in the TFF1 export signature, so that the identifiers can
be printed using their non-overloaded
names like \texttt{real_inv}.

\begin{verbatim}
%   TYPES
tff(tbool, type, bool:$tType).
tff(tfun, type, fn:($tType * $tType) > $tType).
tff(treal, type, real:$tType).
%   CONSTS
tff(cp, type, p : (bool > $o)).
tff(chapp, type, i:!>[A:$tType,B:$tType]: ((fn(A,B) * A) > B)).
tff(cF, type, f:bool).
tff(crealu_inv, type, realu_inv:(real > real)).
tff(cT, type, t:bool).
%   AXIOMS
tff(aEQu_EXT, axiom, ![A : $tType,B : $tType]:
    ![F:fn(A,B),G:fn(A,B)]:(![X:A]:i(A,B,F,X) = i(A,B,G,X) => F = G)).
tff(aBOOLu_CASESu_AX, axiom, ![T:bool]:(T = t | T = f)).
tff(aNOTu_CLAUSESu_WEAKu_conjunct1, axiom, (~ (p(f)) <=> p(t))).
tff(aREALu_INVu_INV, axiom, ![X:real]:realu_inv(realu_inv(X)) = X).
tff(aTRUTH, axiom, p(t)).
tff(conjecture, conjecture, ![X:real,Y:real]:
    (X = Y <=> realu_inv(X) = realu_inv(Y))).
\end{verbatim}
Problems in this format can be already given to the \Why tool, which
can translate them for various SMT solvers and ATP systems, and call
the systems on the translated form. This was initially used both for
ATPs working with the FOF format and for the SMTs. Currently, we only use \Why 
for preparing problems for \Yices, \CVC, and \AltErgo. The
translation to the FOF format was later implemented independently of \Why,
to avoid an additional translation layer for the strongest tools, and in
particular to be able to run the ATPs with different parameters and in
a proof-producing mode. The
procedure is however the same as in \Why, and the resulting FOF
form will be as follows.

\begin{verbatim}
% Goal: !x y. x = y <=> inv x = inv y
fof(aEQu_EXT, axiom, ![A,B]: ![F, G]:
    (![X]: s(B,i(s(fun(A,B),F),s(A,X))) = s(B,i(s(fun(A,B),G),s(A,X)))
    => s(fun(A,B),F) = s(fun(A,B),G))).
fof(aBOOLu_CASESu_AX, axiom,
    ![T]: (s(bool,T) = s(bool,t) | s(bool,T) = s(bool,f))).
fof(aNOTu_CLAUSESu_WEAKu_conjunct1, axiom,
    (~ (p(s(bool,f))) <=> p(s(bool,t)))).
fof(aREALu_INVu_INV, axiom,
    ![X]: s(real,realu_inv(s(real,realu_inv(s(real,X))))) = s(real,X)).
fof(aTRUTH, axiom, p(s(bool,t))).
fof(conjecture, conjecture, ![X, Y]: (s(real,X) = s(real,Y) <=>
    s(real,realu_inv(s(real,X))) = s(real,realu_inv(s(real,Y))))).
\end{verbatim}
This translation uses the (possibly quadratic) tagging of terms with
their types (with ``s'' as the tagging functor), used, e.g., in Hurd's
work.

\subsection{Translations to Higher-Order Formats}

The recently developed TPTP THF standard can be used to encode problems in monomorphic
higher-order logic. This allows experimenting with higher-order ATPs like \LEO~\cite{BenzmullerPTF08}
and \Satallax~\cite{Brown12}, in addition to the standard ATPs working in the first-order formalism.
The translation to THF needs to perform only one
step: monomorphisation.  As explained in~\ref{meson}, this is however a
nontrivial task, and the \MESON tactic approach is already in practice too
exhaustive for problems with many premises.

After developing the TFF1 and FOF translation, some initial
experiments were done to produce a monomorphisation heuristic that
behaves reasonably on problems with many premises.  This heuristic is
now as follows.  The constants that can be used to instantiate the
premises are extracted only once from the goal at the start of the
procedure. Every premise can be instantiated using these goal
constants, but the premises themselves are not further used to grow
this set.  This means that the procedure is even less complete than
\MESON, however the procedure is linear in the number of premises, and
it is therefore possible to use it even with large numbers of advised
premises. In practice, it is rarely the case that a premise can be
instantiated in more than one way. A simple example when this happens is
in the THF
problem\footnote{\url{http://mws.cs.ru.nl/~mptp/hh1/OrigDepsProbs/i/h/00119.p}}
created for the theorem
\th{I_O_ID,}\footnote{\url{http://mws.cs.ru.nl/~mptp/hol-flyspeck/trunk/trivia.html\#I_O_ID}}
where the particular goal and premise (both properties of the identity
function) are as follows:
\begin{verbatim}
Assm: I_THM: !x. I x = x
Goal: I_O_ID: !f. I o f = f /\ f o I = f
\end{verbatim}
The exact types inferred by the standard HOL (Hindley-Milner~\cite{Hindley69}) type inference for the goal are as follows:

$$\forall f : A \to B.\; I_{B \rightarrow B}\ o\ f = f \land f\ o\ I_{A \rightarrow A}  = f $$
Since the identity function appears in the goal both with the type
\texttt{\holthm{A->A}} and with the type \texttt{\holthm{B->B,}}
the following two instances of the premise \texttt{I\_THM}
are created by the THF translation:
\begin{verbatim}
% TYPES
thf(ta, type, a : $tType).
thf(tb, type, b : $tType).
thf(ci0, type, i0 : (a > a)).
thf(ci, type, i : (b > b)).
% AXIOMS
thf(aIu_THMu_monomorphized0, axiom, ![X:a]:((i0 @ X) = X)).
thf(aIu_THMu_monomorphized1, axiom, ![X:b]:((i @ X) = X)).
\end{verbatim}

Finally, while there is no TPTP standard yet for the polymorphic \HOL
logic, this logic is shared by a number of systems in the \HOL family
of ITPs. For the experiments described in~\ref{reproving} \Isabelle is
used in its CASC 2012 THF mode,
but it should be possible to pass the
problems to \Isabelle directly in some (not necessarily TPTP) polymorphic \HOL
encoding. This is has been tried only to a small extent, and is still
future work.

\section{Exporting Theorem Problems for Re-proving with ATPs}
\label{export}

In our earlier initial experiments~\cite{KaliszykU12}, it was found
that the ATP problems created from the calls to the \MESON tactic in
the \HOLLight and \Flyspeck libraries are very easy for the
state-of-the-art ATPs. Some of this easiness might have been caused by
the (generally unsound) merging of different polymorphic versions of
equality used by \MESON into just one standard first-order
equality.\footnote{Note that the typed translation that we use here
  prevents deriving ill-typed equalities~\cite{BlanchettePhd}.} However, after a
manual random inspection it still seemed that the ratio of such unsound
proofs is low, and the \MESON problems are just too easy. That is why only the 
set of problems on the \textit{theorem} level is considered for experiments here. 
The \textit{theorem} level seems to be quite similar in the major ITPs:
\textit{theorem} is typically not corresponding to what mathematicians call
a theorem, but it is rather a self-sufficient lemma with a formal proof
of several to dozens (exceptionally hundreds) lines that can be useful in other
formal proofs and hence should be named and exported. Since the ITP
proofs can be longer (i.e., they can contain a number of \MESON and
other subproblems), proving such theorems fully automatically
is typically a challenge, which makes such problems suitable
for ATP benchmarks, challenges, and competitions.

\subsection{Collecting Theorems and their Dependency Tracking}
In \Mizar/\MPTP and in \Isabelle (done by Blanchette in so far
unpublished work) the ATP problems corresponding to theorems can be
produced by collecting the dependencies (premises) from the proofs (by
suitable tracking mechanisms), and then translating the $Premises
\vdash Theorem$ problem using the methods described in
Section~\ref{translation}.  The recent work by Adams in exporting 
\HOLLight to \HOL Zero~\cite{Adams10} (with cross-verification as the main
motivation) was initially used to obtain the theorem dependencies for
the first experiments with \HOLLight in~\cite{KaliszykU12}, and after that custom theorem-exporting and
dependency-tracking mechanisms were implemented as described below.

\subsubsection{Collecting and Naming of Theorems}

The first issue in implementing such mechanism is to decide what is considered to be a
relevant theorem, and what should be its canonical name. In some ITPs, important statements have labels like
\texttt{lemma}, \texttt{theorem}, \texttt{corollary}, etc. This is not
the case in \HOLLight, which is implemented in the \OCaml
toplevel. This means that every theorem or tactic is just an \OCaml
value. Some of those values are assigned names, while some are only
created on the fly and immediately forgotten.
In the relevant exporting work of Obua~\cite{ObuaSImport}, every occurrence of the \HOLLight command \texttt{prove}
is replaced with a command that additionally records the name of
the stored object. This strategy was used first, and extended to work with
the whole \Flyspeck library by also recording the names for the following commands: \texttt{prove\_by\_refinement},
 \texttt{new\_definition},  \texttt{new\_recursive\_definition},  \texttt{new\_specification}, 
 \texttt{new\_inductive\_definition}, \texttt{define\_type}, and  \texttt{lift\_theorem}.
This purely syntactic replacement method however turned out to be insufficient for a number
of reasons. First, 
this method does not provide information about the scope of names with respect to the \OCaml modules.
Second, it does not provide the information whether a name given to a theorem has been
declared on the top level, or inline inside a function (which makes
such theorem unusable for proving other theorems on the toplevel), 
or even within a function called multiple times with different arguments (in
which case the same name would be assigned to a number of different theorems). Finally,
certain theorems accessible on the top level are created using other \OCaml mechanisms, for example
mapping a decision procedure over a list of terms. Recognizing
syntactically theorems created in this way turned out to be impractical.

That is why a more robust method has been eventually used, based on the
\texttt{update\_database}\footnote{\url{http://mws.cs.ru.nl/~mptp/hol-flyspeck/trunk/Examples/update_database.html}}
 recording functionality by Harrison and
Zumkeller. This code accesses the basic \OCaml data structures and makes it possible to
record the name-value pairs for all \OCaml values of the type \texttt{theorem}
in a given \OCaml state. Thus, it is sufficient to load the whole
\Flyspeck development, and then invoke this recording functionality.

After some initial experiments with ATP re-proving of the translated problems,
this method was however further modified to be able to keep finer track of the
use of theorems that are conjunctions of multiple facts. Such (often large)
conjunctive theorems are used quite frequently in \HOLLight, typically
to package together facts that are likely to jointly provide a useful
method for dealing with certain concepts or certain kinds of
problems. For example the theorem \th{ARITH_EQ}\footnote{\url{http://mws.cs.ru.nl/~mptp/hol-flyspeck/trunk/calc_num.html\#ARITH_EQ}} packages together
ten facts about the equality of numerals as follows:

\begin{holnb}
\holthm{let ARITH_EQ = prove
 (`(!m n. (NUMERAL m = NUMERAL n) <=> (m = n)) /\
   ((_0 = _0) <=> T) /\
   (!n. (BIT0 n = _0) <=> (n = _0)) /\
   (!n. (BIT1 n = _0) <=> F) /\
   (!n. (_0 = BIT0 n) <=> (_0 = n)) /\
   (!n. (_0 = BIT1 n) <=> F) /\
   (!m n. (BIT0 m = BIT0 n) <=> (m = n)) /\
   (!m n. (BIT0 m = BIT1 n) <=> F) /\
   (!m n. (BIT1 m = BIT0 n) <=> F) /\
   (!m n. (BIT1 m = BIT1 n) <=> (m = n))`,
  REWRITE_TAC[NUMERAL; GSYM LE_ANTISYM; ARITH_LE] THEN
  REWRITE_TAC[LET_ANTISYM; LTE_ANTISYM; DENUMERAL LE_0]);;
}\end{holnb}

An even more extreme example is the \texttt{ARITH} theorem which
conjoins together all the basic arithmetic facts (there are 108 of
them in the current version of \HOLLight). The conjuncts of such
theorems are now also named (using a serial numbering of the form
\th{ARITH_EQ_conjunctN}), so that the dependency tracking can
later precisely record which of the conjuncts were used in a
particular proof. This significantly prunes the search space for ATP
re-proving of the theorems that previously depended on the large
conjunctive dependencies, and also
makes the learning data extracted from  dependencies of such theorems more precise.

This method can however result in the introduction of multiple names for a
single theorem (which is just a \HOLLight term of type
\texttt{theorem}). If that happens (for this or other reasons), the
first name that was associated with the theorem during the \Flyspeck
processing is always consistently used, and the other alternative
names are never used.  Such consistency is important for the
performance of the machine learning on the recorded proof
data.\footnote{For example, the \MaLARea system does such
  de-duplication as a useful preprocessing step before learning and
  theorem-proving is started on a large number of related problems.}
The
list of all theorems and their names obtained in this way is saved in
a file, and subsequently used in the dependency extraction and problem
creation passes.

\subsubsection{Dependency Recording}
\label{dependencies}

After the detection and naming of theorems, the recording of proof
dependencies is performed, by processing the whole library again with
a patched version of the \HOLLight kernel. This patched version is
the proof-recording component of the new \HOL-Import~\cite{KaliszykK13},
a mechanism designed to transfer proofs from \HOLLight to \Isabelle/\HOL
in an efficient way allowing the export of big repositories like
\Flyspeck. The code for every \HOL inference step is patched, to record
the newly created theorems. Each theorem is assigned a unique
integer counter, and for every new theorem its dependencies on other
theorems (integers) are recorded and exported to a file.
For every processed theorem it is also checked if it is one of the
theorems named in the previous theorem-naming pass. If so, the association of this
theorem's name to its number is recorded, and again exported to a file.

After this dependency-recording pass, the recorded information is
further processed by an offline program to eliminate all unnamed
dependencies (originating for example from having multiple names for a
single theorem). For every named theorem its dependencies are
inspected, and if a dependency $D$ does not have a name, it is
replaced by its own dependencies (there is no unnamed dependency that
could not be further expanded). This is done recursively, until all
unnamed dependencies are removed. This produces for each named theorem
$T$ a minimal (wrt. the original \HOLLight proof) list of named
theorems that are sufficient to prove $T$. 

The numbering of theorems
respects the order in which the theorems are processed in the
\Flyspeck development. This total ordering is compatible with
(extends) the partial ordering induced by proof dependencies, and for
the experiments conducted here it is assumed to be the
\textit{chronological order} in which the library was developed. 
The dependency information given in
this chronological order for all 16082 named theorems (of
which 1897 are (type) definitions, axioms, or their parts, and their
dependencies are not exported)
 obtained by processing the \Flyspeck 
library\footnote{\Flyspeck
  SVN revision 2887 from 2012-06-26 and \HOLLight SVN revision 146 from
  2012-06-02 are used for all experiments.} 
(and its
\HOLLight pre-requisites) is available
online.\footnote{\url{http://mws.cs.ru.nl/~mptp/hh1/deps.all}} Together
with suitably chosen characterizations of the theorems (see
Section~\ref{features}), this constitutes an interesting new dataset
for machine-learning techniques 
that attempt to predict the most useful premises from the formal library for
proving the next conjecture.

\subsection{The Data Set of ATP Re-proving Problems}

Analogously to \Mizar and \Isabelle, the re-proving ATP problems for
the collected named theorems are finally produced by translating the
$Dependencies \vdash Theorem$ problem to the ATP formalisms using the
methods described in Section~\ref{translation}, together with basic
filtering of dependencies that have trivial first-order content. 1897
of the 16082 named theorems do not have a proof (those are 
definitions and axioms). For all the remaining 14185 named theorems the
corresponding re-proving ATP problems were created, and are available
online\footnote{\url{http://mws.cs.ru.nl/~mptp/hh1/OrigDepsProbs/co_i_f_p_h.tgz}}
in the FOF, THF, and TFF1 formats. These problems are used for the ATP
re-proving experiments described in Section~\ref{reproving}. Smaller
meaningful datasets will likely be created from this large dataset for
ATP/AI competitions such as 
CASC LTB and
Mizar@Turing\footnote{\url{http://www.cs.miami.edu/~tptp/CASC/J6/Design.html\#CompetitionDivisions}}, analogously to the smaller
MPTP2078\footnote{\url{http://wiki.mizar.org/twiki/bin/view/Mizar/MpTP2078}}~\cite{abs-1108-3446}
ATP benchmark created from the ATP-translated \Mizar library (\MML), and the Judgement
Day benchmark~\cite{sledgehammer} created by ATP translation of a
subset of the \Isabelle/\HOL library.

The average, minimum, and maximum sizes of problems in these datasets
are shown in Table~\ref{statsReprove1}, together with the
corresponding statistics for the problems expressed in the original
\HOL formalism.
\begin{table}[bhtp]
\caption{Sizes of the re-proving ATP problems (in numbers of formulas)}
\centering
\begin{tabular}{ccccc}
  \toprule
  Format   & Problems & Average size  & Minimum size & Maximum size  \\\midrule
  HOL      & 14185  & 42.7 & 4 & 510   \\
  FOF      & 14185  & 42.7 & 4     & 510  \\
  TFF1     & 14185 & 71.9  & 10    & 693  \\
  THF      & 14185  & 78.8 & 5     & 1436   \\\bottomrule
\end{tabular}
\label{statsReprove1}
\end{table}
It can be seen that the number of formulas in the translated problems
is typically at most twice the number of the original \HOL formulas,
i.e., the translations are indeed efficient for all the problems. This
was not the case (and became a bottleneck) in the initial experiments using the more prolific
\MESON translation. There is no increase in the number of formulas
when translating from the original \HOL-formulated problem to the FOF
translation. For the TFF1 and THF translation, formulas declaring the
types of the symbols appearing in the problems are added, and for the
THF translation multiple instances of the premises can additionally
appear. Table~\ref{TimesExport1} shows the total times needed
for the various exporting phases run over the whole \Flyspeck as explained above. For completeness, the
time needed to export characterizations of the theorems for external
(e.g., machine-learning) tools is also included (see
Section~\ref{features} for the description of the characterizations
that are used).
\begin{table}[bhtp]
\caption{Times of the exporting phases (total, for the whole \Flyspeck)}
\centering
\begin{tabular}{lc}
  \toprule
  Phase   & Time (minutes)  \\\midrule
  
  standard \Flyspeck loading/verification      & 180       \\
  detection and naming of theorems             & 1          \\
  exporting theorem characterizations          & 5        \\
  dependency recording using patched kernel    & 540    \\ 
  offline post-processing of dependencies      & 10  \\
  creating re-proving ATP problems in FOF      & 34  \\
   creating re-proving ATP problems in TFF1    & 53  \\
    creating re-proving ATP problems in THF    & 32  \\
  total & 855 \\\bottomrule
\end{tabular}
\label{TimesExport1}
\end{table}

\section{Premise Selection}
\label{premiseselection}

Given a large library like \Flyspeck, the interesting ATP/AI task is
to prove new theorems without having to manually select the relevant
premises. In the past decade, a number of premise selection methods
have been developed and experimented with over large theories like
\Mizar/\MML, \Isabelle/\HOL, \SUMO, and \Cyc.  See~\cite{Urban11-ate,KuhlweinLTUH12}
for recent overviews of such methods.

ATP problems of this kind are created for \Mizar/\MML by consistent
translation of the whole \MML to TPTP, and then letting external premise
selection algorithms find the most relevant premises for a given
theorem $t$ from the large set of $t$-allowed premises (typically
those theorems and definitions that were already available when $t$
was being proved, expressed, e.g., as TPTP include files). For
\Isabelle/\Sledgehammer, the default premise selection algorithm is
implemented inside \Isabelle, i.e., it is working on the native
\Isabelle symbols. Only after the \Sledgehammer premise selection
chooses the suitable set of premises, the problem is translated to a
given ATP formalism using one of the several implemented translation
methods. In general, the symbol naming is in \Isabelle consistent only
before the translation is applied, and a particular symbol in two
translated problems can have different meanings.

Both the \Mizar and the \Isabelle approach have some advantages and
disadvantages. Optimizing the translation (or using multiple
translations) as done in \Isabelle can improve the ultimate ATP
performance once the premises have been selected. On the other hand,
if the whole library is not translated in a consistent manner to a
common ATP format such as TPTP, ATP-oriented external premise selection
tools like \SInE cannot be directly used on the whole library. It
could be argued that the \SInE algorithm is relatively close to the
\Sledgehammer premise selection algorithm, and can be easily
implemented inside \Isabelle. However there are useful premise
selection methods for which this is not so straightforward. For
example in the \MaLARea system, evaluation of premises in a large
common pool of finite first-order models is an additional semantic
premise-characterization method that improves the overall precision
quite significantly~\cite{US+08}.\footnote{Recent evidence for 
the usefulness of model-based selection methods is the difference
(64\% vs. 50\% problems solved) between 
the (otherwise quite similar) systems \MaLARea and \PSE (\url{http://www.cs.ru.nl/~kuehlwein/CASC/PS-E.html}) in the 2012 Mizar@Turing competition.}
For such a pool of first-order
models to be useful, the premises have to use
symbols consistently 
also after the translation to first-order logic. Although
various techniques can again be developed to lift this method to the
current \Sledgehammer translation setting, they seem less
straightforward than for example a direct \Isabelle implementation of
\SInE.
This discussion currently applies also to the \HOLLight ATP
translations described in Section~\ref{translation}. For example, the
problem-specific optimization of the arity of symbols described
in~\ref{tff1} will in general cause inconsistency on the symbol level
between the FOF translations of two different \HOLLight problems. 

The procedure implemented for \HOLLight is currently a combination of
the external, internal, learning, and non-learning premise-selection
approaches. This procedure assumes the common ITP situation of a large
library of (also definitional) theorems $T_i$ and their proofs $P_i$
(for definitions the proof is empty), over which a new conjecture has
to be proved. The proofs refer to other theorems, giving rise to a
partial dependency ordering of the theorems extended into their total 
chronological ordering as described for \Flyspeck in~\ref{dependencies}.
For the experiments it will be
assumed that the library was developed in this order.  An overview of
the procedure is as follows, and its details are explained in the
following subsections.
\begin{enumerate}
\item Suitable characterizations (see Section~\ref{features}) of the
  theorems and their proof dependencies are exported from \HOLLight in a simple format.
\item Additional dependency data are obtained by
  running ATPs on the ATP problems created from the \HOLLight proof
  dependencies, i.e., the ATPs are run in the re-proving mode. 
Such data are often smaller and preferable~\cite{KuhlweinU12b}.
These
  data are again exported using the same format as in (1).
\item The (global, first-stage) external premise selectors preprocess (typically train on) 
  the theorem characterizations and the proof dependencies. Multiple characterizations and proof dependencies
  may be used.
\item When a new conjecture is stated in \HOLLight, its
  characterization is extracted and sent to the (pre-trained) first-stage premise selectors.
\item The first-stage premise selectors work as rankers. For a given
  conjecture characterization they produce a ranking of the available
  theorems (premises) according to their (assumed) relevance for the conjecture.
\item The best-ranked premises are used inside \HOLLight to produce
  ATP (FOF, TFF1, THF) problems. Typically several thresholds (8, 32,
  128, 512, etc.) on the number of included premises are used,
  resulting in multiple versions of the ATP problems.
\item The ATPs are called on the problems. Some of the best ATPs
  run in a strategy-scheduling mode combining multiple
  strategies. Some of the strategies always use the \SInE (i.e., local,
  second-stage) premise selection (with different parameters), and
  some other strategies may decide to use \SInE when the ATP problem
  is sufficiently large.
\item[\textit{Loop to improve (2) and (3):}] It is not an uncommon phenomenon that in the data-improving step
  (2) (ATP re-proving from the \HOLLight proof dependencies) an ATP
  proof could not be found for some theorem $T_i$, but an alternative
  proof of $T_i$ can be found from some other theorems preceding $T_i$
  in the chronological order (which guards such alternative proofs
  against cycles). To achieve this, the trained premise selectors can be
  used also on all theorems that are already in the library, and the whole ATP/training process can be iterated several
  times to obtain as many ATP proofs as possible, and better (and 
  differently) trained premise selectors for step (3). 
  This is  the same loop as in \MaLARea.
\end{enumerate}

\subsection{Formula Characterizations Used for Learning}
\label{features}

Given a new conjecture $C$, how do mathematicians decide that certain
previous knowledge will be relevant for proving $C$? The approach
taken in practically all existing premise-selection methods is to
extract from such $C$ a number of suitably defined features, and use
them as the input to the premise selection for $C$.  The most obvious
characterization that already works well in large libraries is the (multi)set
of symbols appearing in the conjecture. This can be further extended
in many interesting ways, using various methods developed, e.g., in
statistical machine translation and web search, but also by methods
specific to the formal mathematical domain.
In this work, characterization of \HOL formulas by all their subterms
(found useful in \MaLARea) was used, and adapted to the typed \HOL
logic. For example, the latest version of the characterization algorithm
would describe the \HOL theorem
\th{DISCRETE_IMP_CLOSED}:\footnote{\url{http://mws.cs.ru.nl/~mptp/hol-flyspeck/trunk/Multivariate/topology.html\#DISCRETE_IMP_CLOSED}}
\begin{holnb}\holthm{
!s:real^N->bool e.
        &0 < e /\ (!x y. x IN s /\ y IN s /\ norm(y - x) < e ==> y = x)
        ==> closed s
}\end{holnb}
by the following set of strings:
\begin{holnb}
"real", "num", "fun", "cart", "bool", "vector_sub", "vector_norm",
"real_of_num", "real_lt", "closed", "_0", "NUMERAL", "IN", "=", "&0",
"&0 < Areal", "0", "Areal", "Areal^A", "Areal^A - Areal^A", 
"Areal^A IN Areal^A->bool", " Areal^A->bool", "_0", "closed Areal^A->bool",
"norm (Areal^A - Areal^A)", "norm (Areal^A - Areal^A) < Areal"
\end{holnb}
This characterization is obtained by:
\begin{enumerate}
\item Normalizing all type variables to just one variable \texttt{A}.
\item Replacing (normalizing) all term variables with their normalized type.
\item Collecting all (normalized) types and their component types recursively.
\item Collecting all (normalized) atomic formulas and their component terms recursively.
\item Including all ``proper'' term and type constructors (logic symbols
  like conjunction 
are filtered
  out). 
\end{enumerate}
In the above example, \texttt{real} is a type constant, \texttt{IN} is
a term constructor, \texttt{Areal\textasciicircum A->bool} is a normalized type,
\texttt{Areal\textasciicircum A} its component type, \texttt{norm
  (Areal\textasciicircum A - Areal\textasciicircum A)
  < Areal} is an atomic formula, and \texttt{Areal\textasciicircum A -
  Areal\textasciicircum A} is its
normalized subterm.

The normalization of variable names is an interesting topic. It is
good if the premise selectors can notice some similarity between two
terms with variables,\footnote{One could require the similarity to
  also handle matching, etc. A simple way how to do it is to generate
  even more features. This is again left to further general research in this area.}
 which is hard (when using strings) if the
variables have different names. On the other hand, total normalization
to just one generic variable name removes also the information that
the variables in a particular subterm were (not) equal. Also,
terms with differently typed variables should be more distant from
each other than those with the same variable types.
In total, four versions of variable normalization were tested:
\begin{itemize}
\item[\texttt{syms0}:] All free and bound variables are given the same
  name \texttt{A0}.
  This encoding is the most liberal, i.e., the
  resulting equality relation on the features is the coarsest one,
  allowing the premise selectors to see many similarities.
\item[\texttt{syms}:] First the free variables are 
numbered consecutively (\texttt{A0}, \texttt{A1}, etc). Then the bound variables are
  named 
with the subsequent numbers. This results in a finer notion of
similarity than in \texttt{syms0}.
\item[\texttt{symst}:] Every variable is renamed to a textual representation
  of its type, for example \texttt{Anum} or \texttt{Areal}. This is
  again finer than \texttt{syms0}, but different from \texttt{syms}. This
  normalization is used in the above example, and also for most of the
  premise selection trainings.
\item[\texttt{symsd}:] In one \texttt{symst} implementation, the
  internal \HOLLight type variable numbering was
accidentally used, thus making most of such term features disjoint
between different theorems. The
performance was lower, but the method produces some unique
solutions and is included in the evaluation.
\end{itemize}
In addition to that, several feature exports included also logic
symbols. 
Various feature characterizations can have different performance on
different datasets, and such characterizations can be also combined
together in interesting ways. This is a rather large research topic
that is left as future work for this newly developed dataset, along
with other large-theory datasets. Just including the features encoding
the validity in finite models will be interesting.

\subsection{Machine Learning of Premise Selection} 

All the currently used first-stage premise selectors are machine
learning algorithms trained in various ways on previous proofs. A
number of machine learning algorithms can be experimented with today,
and in particular kernel-based methods~\cite{abs-1108-3446} and
ensemble methods~\cite{KuhlweinLTUH12} have recently shown quite good
performance on smaller datasets such as MPTP2078.  However, scaling and
tuning such methods to a large corpus like \Flyspeck and to quite a
large number of incremental training and testing experiments is not
straightforward.\footnote{Such scaling up for the large \Mizar library
  is work in progress at the time of writing this article.} That is
why this work so far uses mostly the sparse implementation of a
multiclass naive Bayes classifier provided by the \SNoW system~\cite{Carlson1999}. \SNoW 
can incrementally train and produce predictions on the whole
\Flyspeck library presented in the chronological order in an hour
(and often considerably faster on minimized data). I.e., one new
prediction takes a fraction of a second.
In
addition to that, several other fast incremental (online) learning
algorithms were briefly tried: the Perceptron and Winnow algorithms
provided also by \SNoW, and a custom implementation of the $k$-nearest
neighbor ($k$-NN) algorithm. Only $k$-NN however produced enough additional
prediction power. As already mentioned, the first-stage algorithms are
often complemented by \SInE as a second-stage premise selector when
the ATP problem is written, and that is why some (in particular
\SInE-like) algorithms might look mostly redundant (in the overall ATP
evaluation) when used at the first stage. This is obviously a
consequence of the particular setup used here.

Two kinds of evaluation are possible in this setting and have been
used several times for the \Mizar data: a pure machine-learning
evaluation comparing the predicted premises with the set of known sufficient
premises, and an evaluation that actually runs an ATP on the predicted
premises. While data are available also for the former, in this paper
only the second evaluation is presented, see Section~\ref{advised}.
The main reason for this is that alternative proofs are quite common
in large libraries, and they often obfuscate the link between the pure
machine-learning performance and the final ATP performance. Measuring
the final ATP performance is more costly, however it practically
stopped being a problem with the recent arrival of low-cost
workstations with dozens of CPUs.

At a given point during the library development, the training data
available to the machine learners are the proofs of the previously
proved theorems in the library. A frequently used approach to training
premise selection is to characterize each proof $P_i$ of theorem $T_i$
as a (multi)set of theorems $\{T_{i_1}, ..., T_{i_m} | T_{i_j}\ used\
in\ P_i \}$. 
The training example will then consist
of the input characterization (features) of $T_i$
(see~\ref{features}), and the output characterization (called also
output targets, classes, or labels) will be the (multi)set $\{T_i\}
\cup \{T_{i_1}, ..., T_{i_m} | T_{i_j}\ used\ in\ P_i \}$. Such
training examples can be tuned in various ways. For example the output
theorems may be further recursively expanded with their own
dependencies, the input features could be expanded with the features
of their definitions, various weighting schemes and similarity
clusterings can be tried, etc. This is also mostly left to future
general research in premise-selection learning. 
Once the machine
learner is trained on a particular development state of the library, it is tested
on the next theorem $T$ in the chronological order.  The input
features are extracted from $T$ and given to the trained learner which
then answers with a ranking of the available theorems. This ranking is
given to \HOLLight, which uses it to produce ATP problems for $T$
with varied numbers of the best-ranked premises.

\subsection{Proof Data Used for Learning}
\label{proofdata}

An interesting problem is getting the most useful proof dependencies
for learning. Many of the original \Flyspeck dependencies are clearly
unnecessary for first-order theorem provers. For example the
definition of the $\wedge$ connective (\texttt{AND\_DEF}) is a
dependency of 14122 theorems. Another example are proofs done by
decision procedures, which typically first apply some
normalization steps justified by some lemmas, and then may
perform some standard algorithm, again based on a number of
lemmas. Often only a few of such dependencies are needed (i.e., the
proofs found by decision procedures are often unnecessarily ``complicated'').

Some obviously unhelpful dependencies
were filtered manually, and this was complemented by using also the
data obtained from ATP re-proving (see
Section~\ref{reproving}). \Vampire, \E, and \Z can together re-prove
43.2\% of the \Flyspeck theorems (see Table~\ref{reprovingstats1}),
which is quite a high number, useful for trying to post-process
automatically the remaining dependency data or even to completely
disregard them. The following approaches to combining such ATP and \HOLLight dependencies
were initially tried, and combined with the various characterization
methods to get the training data:
\begin{itemize}
\item[\texttt{minweight} (default):] Always prefer the minimal ATP proof if
  available. On the ATP re-proved theorems collect the statistics
  about how likely a dependency in the \HOLLight proof is really going
  to be used by the ATP proof, and use this likelihood as a weight when ATP
  proof is not available. When the weight is $0$, use (cautiously) a minimal weight
  ($0.001$ or $0.000001$) instead. 
\item[\texttt{nominweight:}] As \texttt{minweight}, but without a
  minimal weight. Totally ignore ATP-irrelevant \HOLLight dependencies.
\item[\texttt{v\_pref} (\texttt{e\_pref, z\_pref}):] Instead of
  using the minimal ATP proof, always prefer the \Vampire (\E, \Z)
  proof. Can be combined with both weighting methods. 
\item[\texttt{symsonly:}] Ignore all proofs. Learn only on
  examples saying that a theorem is good for proving itself, i.e., for
  its feature characterization. The trained system will thus recommend
  theorems similar to the conjecture, but not the dependencies of such
  theorems.
\item[\texttt{atponly:}] Use only the (minimal) ATP proofs for learning. Ignore the
  \HOLLight proofs completely, and construct only the
  \texttt{symsonly} training examples for theorems that have no ATP
  proof. Can be combined with  \texttt{v\_pref}, \texttt{e\_pref, z\_pref}.
\end{itemize}

At some point, a \textit{pseudo-minimization}
procedure started to be applied first to the ATP proofs: each proof is re-run
only with the premises needed for the proof, and if the number of
needed premises decreases, this is repeated until the premise count stabilizes.
Often this further
removes unnecessary premises that appeared in the ATP proof, e.g., by performing
unnecessary rewriting steps. This was later followed by adding
\textit{cross-minimization}: Each proof is re-run (pseudo-minimized) not just by the ATP
that found the proof, but by all ATPs. This can further improve the
training data, and also raise the number of
proofs found by a particular ATP quite considerably, which in turn helps
when proofs by a particular ATP are preferable for learning (see the
\texttt{v\_pref} approach above). Finally, the learning and proving
can boost each other's performance: the proofs obtained by using the
advice of the first-generation premise selectors can be added to the
training data obtained from re-proving, and used to train the second
generation of premise selectors. This process can be iterated, but
only one iteration was done so far (using two different prediction
methods). 

The summary of the training data
obtained by these procedures from the proofs is given in
Table~\ref{trainingpasses}. Each of the ATP-obtained dependency sets
(2--6 in Table~\ref{trainingpasses}) could
be complemented by the  \HOLLight dependencies (1) as described
above, producing differently trained advisors. For example, the best
advising method based on (4) was
only using the ATP proofs for training (no adding of \HOLLight
dependencies when the ATP proof was missing), preferring proofs by \E (\texttt{e\_pref}),
using the \texttt{symst} (types instead of variables) characterizations,
and choosing the best 128 premises. The new ATP proofs found using
this method were added to (4), resulting in the dataset (5). 
The next most complementary advising method to that (measured before
(5) became available) was combining the
ATP dependencies from (2) with the \HOLLight dependencies (1) using the
\texttt{nominweight} and \texttt{v\_pref} techniques, also using 
the \texttt{symst} features, and choosing the
best 512 premises. The new ATP proofs found using
this method were added to (5), resulting in (6). The performance of
various premise selection methods is discussed in Section~\ref{Experiments}.
\begin{table}[bhtp]
\caption{Improving the dependency data used for training premise selection}
\centering
\begin{tabular}{llcccccc>{\bfseries}c>{\bfseries}c}\toprule
\multicolumn{2}{c}{} & \multicolumn{2}{c}{\E} & \multicolumn{2}{c}{\Vampire} &
\multicolumn{2}{c}{\Z} & \multicolumn{2}{c}{\textbf{Union}} \\

 \multicolumn{1}{c}{Nr} & Method  & \multicolumn{1}{c}{thm} & dep  \o&
 \multicolumn{1}{c}{thm} & dep \o&  \multicolumn{1}{c}{thm} & dep \o&  \multicolumn{1}{c}{\textbf{thm}} & dep \o \\\midrule
1&\HOL deps & \multicolumn{6}{c}{}& 14185 & 61.87\\
2&ATP on (1), 300s&   5393 & 4.42& 4700 & 5.15& 4328 & 3.55& 5837 & 3.93\\
3&ATP on (1), 600s$^{*}$&   5655 & 5.80& 5697 & 5.90& 4374 & 3.59& 6161 & 5.00\\
4&(3) minimized &   5646 & 4.52& 5682 & 4.49& 4211 & 3.47& 6104 & 4.35\\
5&  (4) $\cup$  1. advised &   6404 & 4.29& 6308 & 4.17& 5216 & 3.67& 6605 & 4.18\\
6& (5) $\cup$  2. advised &   6848 & 3.90& 6833 & 3.89& 5698 & 3.48& 6998 & 3.81\\\bottomrule
\end{tabular}
{\small
  \begin{description}
  \item[thm:] Number of theorems proved by the given prover.
  \item[dep \o:] The average number of proof dependencies in the
    proofs found by the prover. 
  \item[(1) - \HOL deps:] Dependencies exported from \HOLLight.
  \item[(2),(3):] Dependencies obtained from proofs by ATPs run on \HOL deps.
  \item[(4):] Cross-minimization of (3).
  \item[(5):] Added dependencies from new proofs advised by
    the best learning method  (using (4)).
  \item[(5):] Added dependencies from new proofs advised by
    the best complementary learning method (trained on (2)
    combined with (1)).
  \item[(*):] The 6161 count in (3) is higher than in the final
    900s experiments shown in Table~\ref{reprovingstats1}. This is due
    to a incorrect (cyclic) dependency export for about 60 early \HOLLight
    theorems used for (2)--(4). For training premise selection the
    effect of this error is negligible.
  \end{description}}
\label{trainingpasses}
\end{table}

\subsection{Communication with the Premise Advisors}

There are several modes in which external premise selectors can be
used. The main mode used here for experiments is the \textit{offline
  (batch) mode}. In this mode, the premise selectors are incrementally
trained and tested on the whole library dependencies presented as one
file in the chronological order. \textit{Incremental training/testing}
means that the learning system reads the examples from the file
one by one, for each theorem first producing an advice (ranking of
previous theorems) based only on its features, and only after that
learning on the theorem's dependencies and proceeding to the next
example. The rankings are then used in \HOLLight to produce all ATP
problems in batch mode. This mode is good for experiments, because the
learning data can be analyzed and pre-processed in various ways
described above. All communication is fully file-based.

Another mode is used for \textit{static online advice}. In this mode
an (offline) pre-trained premise selector receives conjecture
characterizations from \HOLLight over a TCP socket, replies in real
time with a ranking of theorems from which \HOLLight produces the ATP
problems and calls the ATPs to solve them. This mode has been
initially implemented  as a simple online service and can already be experimented with by
interested readers.\footnote{The service
  runs at colo12-c703.uibk.ac.at on port 8080, example queries are:\\
  \texttt{echo \textprimstress{}CARD \{3,4\} = 2\textprimstress{} | nc colo12-c703.uibk.ac.at 8080}
  \hspace{1mm}, \\
  \texttt{echo \textprimstress{}(A DIFF B) INTER C = EMPTY <=> (A INTER C)
    SUBSET B\textprimstress{} | nc\\colo12-c703.uibk.ac.at 8080} \hspace{1mm}.}

Finally, in a \textit{dynamic online} mode the premise selector
receives also training data in real time, and updates itself. The
currently used learning systems support this dynamic mode, however, in an
online service this mode of interaction requires some implementation of
access rights, user limits, cloning of developments and services,
etc. This is still future work, close to the recent work on formal mathematical
editors and wikis~\cite{Kaliszyk07,AlamaBMU11}.

\section{Experiments}
\label{Experiments}

The main testing set in all scenarios is constructed from the 14185
\Flyspeck theorems. To be able to explore as many approaches as
possible, a smaller subset of 1419 theorems is often used. This subset
is stable for all such experiments, and it is constructed by taking
every tenth theorem (starting with 0-th) in the chronological ordering.

A number of first- and higher-order ATPs and SMT solvers were tried on
the problems. The most extensively used are \Vampire 2.6 (called also
\V below), modified \E 1.6, and \Z 4.0.  Proofs are important in the
ITP/learning scenarios, so \Z and \E are (unless otherwise noted) run in a
proof-producing mode. In particular for \Z this costs some performance.
\E 1.6 is not run in its standard auto-mode, but in a
strategy-scheduling
wrapper\footnote{\url{https://github.com/JUrban/MPTP2/blob/master/MaLARea/bin/runepar.pl}} 
used by the \MaLARea system in the
Mizar@Turing large-theory competition.  

This wrapper (called either \Epar or just \E in
  the tables below) subsequently tries 14 strategies provided by
  the second author. These strategies were developed on the 1000
  problems allowed for training large-theory AI/ATP systems before the
 Mizar@Turing
competition~\cite{blistr}.
Some of these strategies have become available in \E 1.6 when it was
released after CASC 2012, but \E's auto mode is still tuned for the
TPTP library, and by default it always uses only one ``best'' (heuristically chosen) strategy
on a problem, shunning so far the temptations of strategy scheduling.
\Epar outperforms the old
auto-mode of \E 1.4 by over 20\% on the Mizar@Turing training 
problems, and seems to be competitive with \Vampire 2.6.
Fifteen more systems (or their versions) were tried to a lesser extent
(typically on the 1419-problem subset)
in the experiments. Some of these systems perform very well, and might
be used more extensively later. Sometimes an additional effort is needed
 to make systems really useful; for example, proof/premise output might
 be missing, or additional mapping to a system's constructs needs to
 be done to take full advantage of the system's features. In this work
 such customizations are avoided.
All the systems used are alphabetically listed in Table~\ref{systems}.
\begin{savenotes}
\begin{table}[bhtp]
\caption{Names and descriptions of the systems used in the evaluation}
\centering
\begin{tabular}{lll}
\toprule
System (short) & Format& Description \\\midrule
\AltErgo   &  TFF1 & \AltErgo version 0.94                                                    \\
\CVC       &  TFF1 & \CVC version 2.2                                                          \\
\E 1.6     &  FOF &\E version 1.6pre011 Tiger Hill                                                    \\
\Epar (\E) &  FOF & \E version 1.6pre011 with the Mizar@Turing strategies     \\
\iProver   &  FOF & \iProver version 0.99 (CASC-J6 2012)                                              \\
\Isabelle  &  THF& \Isabelle2012 used in CASC 2012 (without \LEO and \Satallax) \\
\leanCoP   &  FOF & \leanCoP version 2.2 (using eclipse prolog)                                      \\
\LEO-po2   &  THF & \LEO version v1.4.2 in the ``po 2'' proof mode\footnote{For the experiments that produce proof dependencies (useful, e.g., for learning), we have used the (so far experimental) version of \LEO that fully reconstructs the original dependencies (the ``po 2'' option). For the rest of the experiments  (where proofs are not needed), the standard version of \LEO is used. This version is also proof-producing, but some additional work is still needed to extract the original proof dependencies. This work is currently being done by the \LEO developers. The ``po 1'' version outperforms the ``po 2'' version.} (\OCaml -3.11.2)                 \\
\LEO-po1   &  THF & \LEO in the standard ``po 1'' mode (faster)\\
\Metis     &  FOF & \Metis version 2.3 (release 20101019)                                              \\
\Paradox   &  FOF & \Paradox version 4.0, 2010-06-29.                                         \\
\ProverNine&  FOF & \ProverNine (32) version 2009-11A, November 2009.                             \\
\SPASS     &  FOF & \SPASS version 3.5\\
\Satallax  &  THF & \Satallax version 2.6                                                      \\
\Vampire (\V)  &  FOF & \Vampire version 2.6 (revision 1649)                                               \\
\Yices     &  TFF1 & \Yices version 1.0.34                                                    \\
\Z         &  FOF & \Z version 4.0                                                          \\\bottomrule
\end{tabular}
\label{systems}
\end{table}
\end{savenotes}

Unless specified otherwise, all systems are run with 30s time limit on
a 48-core server with AMD Opteron 6174 2.2 GHz CPUs, 320 GB RAM, and
0.5 MB L2 cache per CPU.  Each problem is always assigned one CPU.  In
the tables below, basic statistics are often computed about the
population of the methods used: \textit{Unique} solutions found by
each method, and its \textit{State of the art contribution (SOTAC)} as
defined by CASC.\footnote{For each problem solved by a system, its
  SOTAC for the problem is the inverse of the number of systems that
  solved the problem. A system's overall SOTAC is the average SOTAC
  over the problems it solves.}  A system's CASC-defined SOTAC will be
highest even if the system solved only one problem (which no other
system solved). That is why also the $\mathrm{\Sigma}$-SOTAC value is used: the
sum of a system's SOTAC over all problems attempted.
 These metrics
often indicate how productive it is to add a particular system or its
version to a population of systems.  Often it is interesting to know
the best joint performance when running $N$ methods in
parallel. Finding such a best combination is however an instance of the
classical NP-hard Maximum Coverage problem. While it is often possible
to use SAT solvers to get an optimal solution, a greedy algorithm is
always consistently used to avoid problems when scaling to larger
datasets. This also allows us to present this joint performance as a
simple \textit{greedy (covering) sequence}, i.e., a sequence that 
starts with the best system, and each next system in such sequence is 
the system that greedily adds most solutions to the union of solutions 
of the previous systems in the sequence.

\subsection{Using External ATPs to Prove Theorems from their \HOLLight
Dependencies}
\label{reproving}

Table~\ref{reprovingstats1} shows the results of running \Vampire,
\Epar, \Z, and \Paradox on the 14185 
FOF problems constructed
from the \HOLLight proof dependencies. 
The ATP success rate measured on such problems 
is useful as an upper estimate for the ATP success rate on the
(potentially much larger) problems
where all premises from the whole previous library are allowed to be
used. This success rate
can be used later to evaluate the performance of the algorithms that
select a smaller number of the most relevant premises.
The time limit for \Vampire, \Epar and \Z was relatively high (900s),
because particularly \Vampire benefits from higher time limits
(compare with Table~\ref{allreproving1}) and the ATP proofs found by
re-proving turn out to be more useful for training premise selectors
than the original \HOLLight dependencies (see
Section~\ref{advised}). \Paradox was run for 30 seconds to get some
measure of the incompleteness of the FOF translation. The systems in
Table~\ref{reprovingstats1} are already ordered using the
\textit{greedy covering sequence}, i.e., the joint performance of the top two
systems is 41.9\%, etc. The counter-satisfiability detected
by \Paradox is not by default included in the \textit{greedy
  sequence}, since its goal is to find the strongest combination of
proof-finding systems. The \Paradox results are however included in
the SOTAC and Unique columns.
\begin{table}[bhtp]
\caption{\Epar, \V, \Z re-proving with 900s and \Paradox with 30s
  (14185 problems)}
\centering
\begin{tabular}{lccccccc}\toprule
Prover & Theorem (\%) & Unique & SOTAC & $\mathrm{\Sigma}$-SOTAC & CounterSat (\%) & Greedy (\%) \\\midrule
\Vampire & 5641 (39.7) & 218 & 0.403 & 2273.58 & 0 ( 0.0) & 5641 (39.7) \\
\Epar & 5595 (39.4) & 194 & 0.400 & 2238.58 & 0 ( 0.0) & 5949 (41.9) \\
\Z & 4375 (30.8) & 193 & 0.372 & 1629.08 & 2 ( 0.0) & 6142 (43.2) \\
\Paradox & 5 ( 0.0) & 0 & 0.998 & 2612.75 & 2614 (18.4) & 6142 (43.2) \\
any & 6142 (43.2) & & & & 2614 (18.4) & \\\bottomrule
\end{tabular}
{\small
  \begin{description}
  \item[Theorem (\%):] Number and percentage of theorems proved by a system.
  \item[Unique:] Number of theorems proved only by this system.
  \item[SOTAC, $\mathrm{\Sigma}$-SOTAC:] See the explanatory text for these metrics.
  \item[CounterSat:] Number of problems found counter-satisfiable (unprovable) by this system.
  \item[Greedy (\%):] The joint coverage of all the previous systems and this one ordered in a greedy sequence (see the text).
  \end{description}}
\label{reprovingstats1}
\end{table}

Table~\ref{reprovingstats2} shows these results restricted to the
1419-problem subset.  This provides some measure of the statistical
error encountered when testing systems on the smaller problem set, and
also a comparison of the systems' performance under high and low time
limits used in Table~\ref{allreproving1}.
\begin{table}[bhtp]
\caption{ATP re-proving with 900s time limit on 10\% (1419) problems}
\centering
\begin{tabular}{lccccccc}\toprule
Prover & Theorem (\%) & Unique & SOTAC & $\mathrm{\Sigma}$-SOTAC & CounterSat (\%) & Greedy (\%) \\\midrule
\Vampire & 577 (40.6) & 19 & 0.403 & 232.25 & 0 ( 0.0) & 577 (40.6) \\
\Epar & 572 (40.3) & 23 & 0.405 & 231.75 & 0 ( 0.0) & 608 (42.8) \\
\Z & 436 (30.7) & 17 & 0.369 & 160.75 & 0 ( 0.0) & 625 (44.0) \\
\Paradox & 1 ( 0.0) & 0 & 0.997 & 257.25 & 257 (18.1) & 625 (44.0) \\
any & 625 (44.0) & & & & 257 (18.1) & \\\bottomrule
\end{tabular}

\label{reprovingstats2}
\end{table}

Table~\ref{allreproving1} shows all tested systems on the 1419-problem
subset, ordered by their absolute performance, and
Table~\ref{allreproving2} shows the corresponding greedy ordering. The
tested systems include also SMT solvers that use the TFF1 encoding and
higher-order provers using the THF encoding. This is why it is no
longer possible to aggregate the counter-satisfiability results
(particularly found by \Paradox) with the theoremhood results, and all
the derived statistics are only computed using the Theorem column.
While \Vampire does well with high time limits in
Table~\ref{reprovingstats1}, it is outperformed by \Z and \E-based
systems (\Epar, \E 1.6, \LEO-po1) when using only 30 seconds (which
seem more appropriate for interactive tools than 300 or even 900
seconds). This suggests that the strategy scheduling in \Vampire might
benefit from further tuning on the \Flyspeck data.  \Z is not run in
the proof-producing mode in this experiment, which improves its
performance considerably.
It is not very
surprising (but still evidence of solid integration work) that
\Isabelle performs best, as it already combines a number of other
systems; see its CASC 2012
description\footnote{\url{www.cs.miami.edu/~tptp/CASC/J6/SystemDescriptions.html\#Isabelle---2012}}
for details. An initial glimpse at \Isabelle's unique solutions also
shows that 75\% of them are found by the recent \Isabelle-specific
additions (such as hard sorts) to \SPASS~\cite{BlanchettePWW12} and its tighter integration
with \Isabelle.
This is an evidence
that pushing such domain knowledge inside ATPs (as done recently also
with the \MaLeCoP prototype~\cite{UrbanVS11}) might be quite
rewarding.  The joint performance of all systems tested is
50.2\% when \Isabelle is included, and 47.4\% when only the base
systems are allowed. 
This is quite encouraging, and for example the
counter-satisfiability results suggest that additional performance
could be gained by further (possibly heuristic/learning) work on
alternative translations. Pragmatically, the joint re-proving performance also
tells us that
when used in the \MESON-tactic mode with premises explicitly provided
by the users, a parallel 9-CPU machine running the nine systems from
Table~\ref{allreproving2} will within 30 seconds (of real time) prove half of the
\Flyspeck theorems without any further interaction.

\begin{table}[bhtp]
\caption{All ATP re-proving with 30s time limit on 10\% of problems}
\centering
\begin{tabular}{lcccccc}\toprule
Prover & Theorem (\%) & Unique & SOTAC & $\Sigma$-SOTAC & CounterSat (\%) & Processed \\\midrule
\Isabelle & 587 (41.3) & 39 & 0.201 & 118.09 & 0 ( 0.0) & 1419 \\
\Epar & 545 (38.4) & 9 & 0.131 & 71.18 & 0 ( 0.0) & 1419 \\
\Z & 513 (36.1) & 17 & 0.149 & 76.49 & 0 ( 0.0) & 1419 \\
\E 1.6 & 463 (32.6) & 0 & 0.101 & 46.69 & 0 ( 0.0) & 1419 \\
\LEO-po1 & 441 (31.0) & 1 & 0.106 & 46.85 & 0 ( 0.0) & 1419 \\
\Vampire & 434 (30.5) & 3 & 0.107 & 46.44 & 0 ( 0.0) & 1419 \\
\CVC & 411 (28.9) & 4 & 0.111 & 45.76 & 0 ( 0.0) & 1419 \\
\Satallax & 383 (26.9) & 7 & 0.130 & 49.69 & 1 ( 0.0) & 1419 \\
\Yices & 360 (25.3) & 0 & 0.097 & 35.06 & 0 ( 0.0) & 1419 \\
\iProver & 348 (24.5) & 0 & 0.088 & 30.50 & 9 ( 0.6) & 1419 \\
\ProverNine & 345 (24.3) & 0 & 0.087 & 30.07 & 0 ( 0.0) & 1419 \\
\Metis & 331 (23.3) & 0 & 0.085 & 28.23 & 0 ( 0.0) & 1419 \\
\SPASS & 326 (22.9) & 0 & 0.081 & 26.46 & 0 ( 0.0) & 1419 \\
\leanCoP & 305 (21.4) & 1 & 0.092 & 27.96 & 0 ( 0.0) & 1419 \\
\AltErgo & 281 (19.8) & 1 & 0.100 & 28.14 & 0 ( 0.0) & 1419 \\
\LEO-po2 & 53 ( 3.7) & 0 & 0.082 & 4.34 & 0 ( 0.0) & 1419 \\
\Paradox & 1 ( 0.0) & 0 & 0.059 & 0.06 & 259 (18.2) & 1419 \\
any & 712 (50.1) & & & & 259 (18.2) & 1419 \\\bottomrule
\end{tabular}
\label{allreproving1}
\end{table}

\begin{table}[bhtp]
\caption{Greedy sequence for Table~\ref{allreproving1} (with and
  without \Isabelle)}
\centering
\begin{tabular}{ccccccccccc}
\multicolumn{1}{c}{Prover}\\\multicolumn{1}{c}{}&
\multicolumn{1}{c}{\begin{rotate}{45}\Isabelle\end{rotate}}&
\multicolumn{1}{c}{\begin{rotate}{45}\Z\end{rotate}}&
\multicolumn{1}{c}{\begin{rotate}{45}\Epar\end{rotate}}&
\multicolumn{1}{c}{\begin{rotate}{45}\Satallax\end{rotate}}&
\multicolumn{1}{c}{\begin{rotate}{45}\CVC\end{rotate}}&
\multicolumn{1}{c}{\begin{rotate}{45}\Vampire\end{rotate}}&
\multicolumn{1}{c}{\begin{rotate}{45}\LEO-po1\end{rotate}}&
\multicolumn{1}{c}{\begin{rotate}{45}\leanCoP\end{rotate}}&
\multicolumn{1}{c}{\begin{rotate}{45}\AltErgo\end{rotate}}\\ \toprule
 Sum \%&
 41.3&
 46.9&
 48.8&
 49.3&
 49.6&
 49.9&
 50.0&
 50.1&
 50.1\\
 Sum&
 587&
 666&
 693&
 700&
 705&
 709&
 710&
 711&
 712\\\bottomrule
\end{tabular}
\\[1cm]
\begin{tabular}{ccccccccccc}
\multicolumn{1}{c}{Prover}\\\multicolumn{1}{c}{}&
\multicolumn{1}{c}{\begin{rotate}{45}\Epar\end{rotate}}&
\multicolumn{1}{c}{\begin{rotate}{45}\Z\end{rotate}}&
\multicolumn{1}{c}{\begin{rotate}{45}\Satallax\end{rotate}}&
\multicolumn{1}{c}{\begin{rotate}{45}\Vampire\end{rotate}}&
\multicolumn{1}{c}{\begin{rotate}{45}\LEO-po1\end{rotate}}&
\multicolumn{1}{c}{\begin{rotate}{45}\CVC\end{rotate}}&
\multicolumn{1}{c}{\begin{rotate}{45}\AltErgo\end{rotate}}&
\multicolumn{1}{c}{\begin{rotate}{45}\Yices\end{rotate}}&
\multicolumn{1}{c}{\begin{rotate}{45}\leanCoP\end{rotate}}\\\toprule 
 Sum \%&
 38.4&
 42.7&
 45.3&
 46.0&
 46.6&
 47.1&
 47.2&
 47.3&
 47.4\\
 Sum&
 545&
 607&
 644&
 654&
 662&
 669&
 671&
 672&
 673\\\bottomrule
\end{tabular}
\label{allreproving2}
\end{table}

\subsection{Using External ATPs to Prove Theorems with Premise Selection}
\label{advised}
As described in Section~\ref{premiseselection}, there are a number of
various approaches and parameters influencing the training of the
premise selectors. These parameters were gradually (but not
exhaustively) explored, typically on the 1419-problem subset. Several
times the underlying training data changed quite significantly as a
result of the data-improving passes described in
Section~\ref{proofdata}. Some of these passes were evaluating the best prediction methods developed
so far on all 14185 problems. All the experiments were limited to \Vampire,
\Epar, and \Z . For most of the experiments (and unless otherwise
noted) the first-stage premise selection is used to create problems
with 8, 32, 128, and 512 premises. This slicing (i.e., taking the first $N$ premises) can be later fine-tuned, as
done below in Section~\ref{slices} for the best premise selection method.

Table~\ref{firstround} shows an initial evaluation of 16 different learning
combinations trained on ATP proofs obtained in 300s, complemented by
the \HOLLight proof dependencies (the second and first pass in
Table~\ref{trainingpasses}). The two exceptions are the
\texttt{symst+symsonly} combination, which ignores all proofs, and the
\texttt{syms+old+000001} combination, which uses older ATP-re-proving
data obtained by running each ATP only for 30s (about 700 ATP proofs
less). Each row in Table~\ref{firstround} is a union of twelve 30s ATP
runs: \Vampire, \Epar, and \Z used on the 8, 32, 128, and 512
slices. 
After this initial evaluation, the
\texttt{symst} (types instead of variables) characterization was
preferred, trivial symbols were always pruned out, and Winnow and
Perceptron were left behind. It is of course possible that some of these
methods are useful as a complement of better methods.
Preferring \Vampire proofs helps the learning a bit, for reasons that
are not yet understood.
To get the joint 39.5\% performance, in
general 192-fold ($= 16\times12$) parallelization is needed. This number could be
reduced, but first better training methods were considered.

\begin{table}[bhtp]
\caption{16 premise selection methods trained on ATP proofs
  complemented with \HOL proofs}
\centering

\begin{tabular}{>{\ttfamily}lccccc}\toprule
Method 				& Theorem (\%) & Unique & SOTAC & $\mathrm{\Sigma}$-SOTAC & Processed \\\midrule
B+symst+v\_pref+nominweight	& 418 (29.4) & 2 & 0.093 & 38.98 & 1419 \\ 
B+symst				& 410 (28.8) & 0 & 0.089 & 36.65 & 1419 \\ 
B+syms0 			& 406 (28.6) & 0 & 0.084 & 33.99 & 1419 \\ 
B+symst+triv			& 405 (28.5) & 3 & 0.094 & 37.98 & 1419 \\ 
B+syms				& 402 (28.3) & 0 & 0.083 & 33.48 & 1419 \\ 
B+syms+triv+nominweight		& 397 (27.9) & 1 & 0.083 & 32.98 & 1419 \\ 
B+syms0+triv 			& 397 (27.9) & 0 & 0.078 & 30.82 & 1419 \\ 
B+syms+triv+v\_pref 		& 396 (27.9) & 0 & 0.081 & 31.99 & 1419 \\ 
B+syms+triv			& 393 (27.6) & 0 & 0.077 & 30.13 & 1419 \\ 
B+syms+triv+z\_pref		& 392 (27.6) & 1 & 0.082 & 32.09 & 1419 \\ 
B+syms+triv+e\_pref		& 392 (27.6) & 0 & 0.078 & 30.45 & 1419 \\ 
B+symsd				& 382 (26.9) & 3 & 0.097 & 37.08 & 1419 \\ 
B+syms+old+000001		& 376 (26.4) & 14 & 0.129 & 48.34 & 1419 \\
B+symst+symsonly		& 302 (21.2) & 17 & 0.141 & 42.55 & 1419 \\
W+symst				& 251 (17.6) & 9 & 0.141 & 35.37 & 1419 \\ 
P+symst				& 195 (13.7) & 6 & 0.144 & 28.13 & 1419 \\ 
any                             & 561 (39.5) & & & & 1419 \\\bottomrule
\end{tabular}
{\small
  \begin{description}
  \item[\texttt{B,W,P:}] Naive Bayes, Winnow, Perceptron.
  \item[\texttt{triv:}] Logical (trivial) symbols like conjunction are included.
  \item[\texttt{old+000001:}] Using older 30-second ATP data, and a minimal
    weight of 0.000001 for irrelevant \HOL dependencies (the default
    weight was 0.001). 
  \end{description}}

\label{firstround}
\end{table}

\subsubsection{Further Premise Selection Improvements}

Complementing the ATP dependency data with the (possibly discounted)
\HOLLight dependencies seems to be a plausible method. Even if the
\HOLLight dependencies are very redundant, the redundancies should be
weighted down by the information learned from the large number of ATP
proofs, and the remaining \HOLLight dependencies should be in general
more useful than no information at all. A possible explanation of why
this approach might still be quite suboptimal (in the ATP setting) is
that the \HOLLight proofs are often not a good guidance for the ATP
proofs, and may push the machine learners in a direction that is
ATP-infeasible. A small hint that this might be the case is the good
performance of the \texttt{nominweight} method in
Table~\ref{firstround}. This method completely ignores all \HOLLight
dependencies that were never used in previous ATP proofs. This
suggested to test the more radical \texttt{atponly} approach, in which
only the ATP proofs are used for training. This approach improved the
best method from 29.4\% to 31.9\%, and added 25 newly solved problems
(1.8\%) to those solved by the 16 methods in
Table~\ref{firstround}. These results motivated further work on
getting as many (and as minimal) ATP proofs as possible, producing 
the methods tagged as \texttt{m10, m10u} and \texttt{m10u2} 
in the tables below. These methods were trained on the proofs obtained
by 10-minute (hence \texttt{m10}) ATP runs that were further upgraded
by the advised proofs as described in Section~\ref{proofdata}. The
best \texttt{m10u2} method raised the performance by further 0.5\%,
and the learning on the advised proofs in different ways made these
methods again quite orthogonal to the previous ones.

Even though Winnow and Perceptron performed poorly (as expected from
earlier unpublished experiments with \MML), they added some
new solutions. This motivated one simple additional experiment with
the classic $k$-nearest
neighbor ($k$-NN) learner, which computes for a new example
(conjecture) the $k$ nearest (in a given feature distance)
previous examples and ranks premises by their frequency in these examples.
This is a fast (``lazy'' and trivially incremental) learning method that can be
easily parameterized and might for 
some parameters behave quite differently from naive Bayes. For large
datasets a basic implementation gets slow in the evaluation phase,
but on the \Flyspeck dataset this was not yet a problem and full
training/evaluation processing took about the same time as naive Bayes.
Table~\ref{knn1} shows the performance of three differently
parameterized $k$-NN instances, and Table~\ref{knn2} shows 8 different
$k$-NN-based methods that together prove 29\% of the problems. As
expected, $k$-NN performs worse than naive Bayes, but much better than
Winnow and Perceptron. The $160$-NN and $40$-NN methods indeed produce
somewhat different solutions, and they are sufficiently orthogonal to
the previous methods and both contribute to the
performance of the final best mix of 14 prediction/ATP methods.

\begin{table}[bhtp]
\caption{Three instances of $k$-nearest neighbor on the
  1419-problem subset}
\begin{center}
\begin{tabular}{*>{\ttfamily}l^c^c^c^c^c}\toprule
Method & Theorem (\%) & Unique & SOTAC & $\mathrm{\Sigma}$-SOTAC &  Processed \\\midrule
KNN160+m10u+atponly & 391 (27.5) & 84 & 0.512 & 200.33             & 1419 \\
KNN40+m10u+atponly & 330 (23.2) & 10 &  0.403 & 132.83 & 1419 \\
KNN10+m10u+atponly & 244 (17.1) & 6 &   0.360 & 87.83  & 1419 \\
any & 421 (29.6) & & & & 1419 \\\bottomrule
\end{tabular}
\end{center}
\label{knn1}
\end{table}

\begin{table}[bhtp]
\caption{Greedy sequence using $k$-NN-based premise selectors}
\begin{center}
\begin{tabular}{*c^c^c^c^c}\toprule

\rowstyle{\ttfamily}
Method& KNN160+512e  & KNN40+32e    & KNN160+512v  & KNN40+512e   \\\midrule
Sum \% & 21.7 & 25.5 & 26.7 & 27.4 \\   
Sum  & 309 & 363 & 379 & 390 \\\midrule 
\rowstyle{\ttfamily}
...& KNN160+128z  & KNN160+128e  & KNN10+512v   & KNN10+512e \\\midrule
... & 28.1 & 28.4 & 28.8 & 29.0 \\   
...& 399 & 404 & 409 & 412 \\\bottomrule
\end{tabular}
\end{center}
\label{knn2}
\end{table}

\subsubsection{Performance of different premise slices}
\label{slices}
\begin{figure}[thb]
\caption{Performance of different premise slices (in \% of the
  1419 problems)}
\begin{center}
\includegraphics[width=12cm]{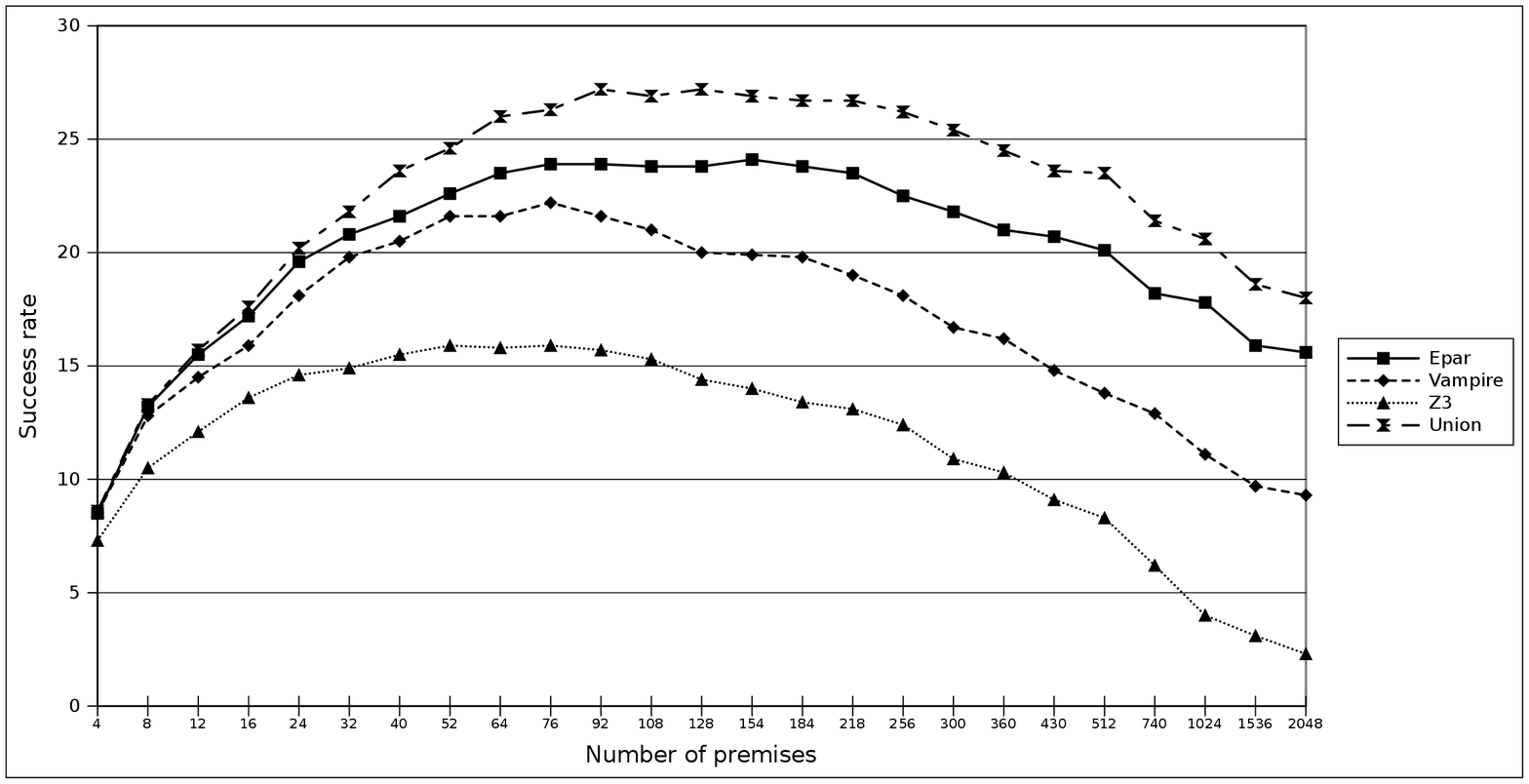}
\end{center}
\label{premiseslices}
\end{figure}

Fig.~\ref{premiseslices} shows how the ATP performance changes when
using different numbers of the best-ranked premises. This is again
evaluated in 30 seconds on the 1419-problem subset, i.e., \Vampire's
performance is likely to be better (compared to \Epar) if a higher
time limit was used. To a certain extent this graph serves also as a
comparison of the first-stage premise selection (in this case naive
Bayes trained on the minimized proofs) with the second-stage premise
selection (various \SInE strategies tried by the ATPs). \Z has no
second-stage premise selection, and after 250 premises the performance
drops quite quickly (12.4\% with 256 premises vs. 6.2\% with 740
premises). For \Vampire this drop is more moderate (18.1\%
vs. 12.9\%). \Epar stays over 20\% with 512 premises, and drops
only to 15.6\% with 2048 premises. Thus, 512 premises seems to be the
current ``margin of error'' for the first-stage premise selection that
can be (at least to some extent) offset by using \SInE at the second
stage. 

Table~\ref{jointbestslices} shows for this premise selection method
the joint performance (in greedy steps) of all premise
slicings, when for each slice the union of all ATPs' solutions is
taken. Only 17 slices are necessary (when using the greedy approach);
the remaining 8 slices do not contribute more solutions. 
In general, this union would take $3*17 = 51$ ATP runs, however only
28 ATP runs are actually required to achieve the maximum 36.4\%
performance. These runs are not shown in full here, and instead only
the first 14 runs that yield 35\% are shown in Table~\ref{bestslicesatp}.
Assuming a 14-CPU server, 35\% is thus the 30-second performance when
using only one (the best) premise selection method.

\begin{table}[bhtp]
\caption{Greedy covering sequence for \texttt{m10u2} slices (joining ATPs)}
\begin{center}
\begin{tabular}{*c^c^c^c^c^c^c^c^c^c^c}\toprule

\rowstyle{\bfseries}
Method& 128  & 1024 & 40   & 256  & 64   & 24   & 740  & 430& ... \\\midrule  
Sum \%& 27.2& 30.2& 32.9& 33.9& 34.5& 34.9& 35.3& 35.5& ...
\\
 Sum& 386& 429& 467& 482& 490& 496& 501& 504& ...
\\\midrule 

\rowstyle{\bfseries} ... &
92  & 2048  & 184  & 218  & 154  & 52  & 32  & 12 & 4\\\midrule  
...& 35.7& 35.8& 36.0& 36.0& 36.1& 36.2& 36.2& 36.3& 36.4\\
 ...& 507& 509& 511& 512& 513& 514& 515& 516& 517
\\\bottomrule
\end{tabular}
\end{center}
\label{jointbestslices}
\end{table}

\begin{table}[bhtp]
\caption{Greedy covering sequence for \texttt{m10u2} limited to 14 slicing/ATP methods}
\begin{center}
\begin{tabular}{*c^c^c^c^c^c^c^c^c}\toprule

\rowstyle{\bfseries}
Method& 154e& 52v& 1024e& 16e& 300v& 92e& 64z \\\midrule  
Sum \%& 24.1& 27.4& 29.8& 31.1& 32.0& 32.6& 33.1 \\
 Sum& 343& 390& 423& 442& 455& 464& 471 \\\midrule 

\rowstyle{\bfseries}
... & 256z& 256e& 184v& 32v& 2048e& 128v& 24z\\\midrule  
... & 
33.5& 33.8& 34.1& 34.3& 34.6& 34.8& 35.0 \\
... & 
476& 480& 484& 488& 491& 494& 497 \\\bottomrule
\end{tabular}
\end{center}
\label{bestslicesatp}
\end{table}

\subsubsection{The Final Combination and Higher Time Limits}

It is clear that the whole learning/ATP AI system can (and will) be
(self-)improved in various interesting ways and for long time.\footnote{In
  2008, new proofs were still discovered after a month
  of running \MaLARea on the whole \MML. Analyzing the proofs and improving
  such AI over an interesting corpus gets addictive.}  When the number of small-scale
evaluations reached several hundred and the main initial issues
seemed corrected, an overall evaluation of the (greedily) best combination of 14
methods was done on the whole set of 14185 \Flyspeck problems using a
300s time limit. These 14 methods together prove 39\% of the theorems
when given 30 seconds in parallel (see Table~\ref{stats0}), which is also how they
are run in the online service.
The large scale evaluation is shown in Table~\ref{stats1} and Table~\ref{stats2}.
Table~\ref{stats1} sorts the methods 
by their 300s performance, and Table~\ref{stats2} computes the
corresponding greedy covering sequence.
Comparison with Table~\ref{stats0} shows that raising the CPU time to 300s
helps the individual methods (2.7\% for the best one), but not so much the
final combination (only 0.3\% improvement).

\begin{table}[bhtp]
\caption{The top 14 methods in the greedy sequence for 30s small-scale
runs}
\centering
\begin{tabular}{>{\ttfamily}lcc}\toprule
Prover & Sum \% & Sum \\\midrule
B+symst+m10u2+atponly+154e	& 24.1 &    343 \\
KN40+symst+m10u+atponly+32e	& 28.6 &    407 \\
B+symst+m10u2+atponly+1024e	& 31.2 &    443 \\
B+v_pref+triv+128e		& 33.1 &    471 \\
B+symst+m10u2+atponly+92v 	& 34.4 &    489 \\
W+symst+128e			& 35.3 &    501 \\
B+symst+m10u+atponly+32z	& 35.9 &    510 \\
B+syms0+triv+512e		& 36.5 &    519 \\
B+all000001+old+triv+128v	& 37.2 &    528 \\
KN160+symst+m10u+atponly+512z& 37.6 &    534    \\
W+symst+32z			& 37.9 &    539 \\
B+symst+m10u2+atponly+128e	& 38.3 &    544 \\
B+symst+m10+vs+pref+atponly+128z& 38.6 & 549    \\
B+symst+32z			& 39.0 &    554 \\\bottomrule
\end{tabular}
\label{stats0}
\end{table}

\begin{table}[bhtp]
\caption{The top 14 methods from Table~\ref{stats0} evaluated in 300s on all 14185 problems}
\centering
\begin{tabular}{>{\ttfamily}lcccc}\toprule
Prover&Theorem (\%)&Unique&SOTAC&$\mathrm{\Sigma}$-SOTAC\\\midrule
B+symst+m10u2+atponly+154e           & 3810 (26.8) & 33 & 0.157 & 597.89 \\
B+symst+m10u2+atponly+128e           & 3799 (26.7) & 25 & 0.153 & 580.70 \\
B+symst+m10u2+atponly+92v            & 3740 (26.3) & 95 & 0.167 & 623.03 \\
B+symst+m10u2+atponly+1024e          & 3280 (23.1) & 206 & 0.198 & 649.52 \\
B+syms0+triv+512e                    & 3239 (22.8) & 101 & 0.170 & 551.90  \\
B+v_pref+triv+128e                   & 2814 (19.8) & 36 & 0.143 & 401.27  \\
B+all000001+old+triv+128v            & 2475 (17.4) & 50 & 0.149 & 367.86  \\
KN40+symst+m10u+atponly+32e & 2417 (17.0) & 78 & 0.160 & 386.90 \\
B+symst+m10+v_pref+atponly+128z & 2257 (15.9) & 33 & 0.138 & 311.74\\
B+symst+m10u+atponly+32z     & 2191 (15.4) & 43 & 0.139 & 304.77 \\
KN160+symst+m10u+atponly+512z & 1872 (13.1) & 37 & 0.145 & 270.58  \\
W+symst+128e                         & 1704 (12.0) & 56 & 0.164 & 279.34 \\
B+symst+32z                          & 1408 ( 9.9) & 16 & 0.118 & 166.09  \\
W+symst+32z                          & 711 ( 5.0) & 9 & 0.124 & 88.40  \\
any & 5580 (39.3) & & & \\\bottomrule
\end{tabular}

\label{stats1}
\end{table}

\begin{table}[bhtp]
\caption{The greedy sequence for Table~\ref{stats1} (300s runs on all problems)}
\centering
\begin{tabular}{>{\ttfamily}lcc}\toprule
Prover & Sum \% & Sum \\\midrule
B+symst+m10u2+atponly+154e       & 26.8 & 3810 \\
B+symst+m10u2+atponly+1024e      & 30.1 & 4273 \\
B+symst+m10u2+atponly+92v        & 33.0 & 4686 \\
KN40+symst+m10u+atponly+32e      & 34.8 & 4938 \\
B+syms0+triv+512e                & 36.2 & 5148 \\
B+all000001+old+triv+128v        & 36.9 & 5247 \\
B+symst+m10u+atponly+32z         & 37.5 & 5332 \\
W+symst+128e                     & 38.1 & 5411 \\
KN160+symst+m10u+atponly+512z    & 38.4 & 5454 \\
B+v_pref+triv+128e               & 38.7 & 5492 \\
B+symst+m10+v_pref+atponly+128z & 38.9 & 5528 \\
B+symst+m10u2+atponly+128e       & 39.1 & 5553 \\
B+symst+32z                      & 39.2 & 5571 \\
W+symst+32z                      & 39.3 & 5580 \\\bottomrule
\end{tabular}
\label{stats2}
\end{table}

\subsection{Union of Everything}
\label{union}
Tables~\ref{union0} shows the ``union of everything'',
i.e., the union of problems (limited to the 1419-problem subset) that could either be proved by an ATP from the \HOLLight dependencies or 
by the premise selection methods. Together with Table~\ref{allreproving1} and Table~\ref{allreproving2} this also shows how much the ATP proofs obtained by 
premise selection methods complement the ATP proofs based only on the \HOLLight dependencies.
The methods' running times are not
comparable: the re-proving used 30s for each system, while the data
for advised methods are aggregated across  \E, \Vampire and \Z, 
and across the four premise slicing methods. This means that they in
general run  in $12\times30$ seconds (although typically only one or two
slices are needed for the final joint performance).
The number of \Flyspeck theorems that were proved by any of the
many experiments conducted is thus 56.5\% when \Isabelle is considered, and
54.7\% otherwise.

\begin{table}[bhtp]
\begin{small}
\caption{Covering sequence (without and with \Isabelle) for all
  methods used}
\begin{tabular}{>{\ttfamily}lcc}\toprule
Prover & Sum \% & Sum \\\midrule
\Epar & 38.4 & 545 \\
B+syms+triv & 44.7 & 635 \\
\Z & 48.2 & 684 \\
\Satallax & 50.7 & 720 \\
B+v\_pref+atponly & 52.1 & 740 \\
\LEO-po1 & 52.6 & 747 \\
\CVC & 53.0 & 753 \\
B+symsonly+triv & 53.4 & 758 \\
B+symst+m10+nominwght & 53.6 & 762 \\
KN40+symst+m10u+atponly & 53.9 & 765 \\
B+syms0 & 54.1 & 768 \\
W+symst & 54.2 & 770 \\
B+symst+m10u2+atponly & 54.4 & 772 \\
\Vampire & 54.4 & 773 \\
\leanCoP & 54.5 & 774 \\
B+symst & 54.6 & 775 \\
B+symst+triv & 54.6 & 776 \\
\AltErgo & 54.7 & 777 \\\bottomrule
\end{tabular}
\quad
\begin{tabular}{>{\ttfamily}lcc}\toprule
Prover & Sum \% & Sum \\\midrule
\Isabelle & 41.3 & 587 \\
B+symst+m10u2+atponly & 48.4 & 687 \\
\Z & 52.0 & 738 \\
\Epar & 53.0 & 753 \\
B+syms & 54.0 & 767 \\
B+symsonly+triv & 54.5 & 774 \\
\Satallax & 54.9 & 780 \\
B+symst+m10+nominwght & 55.3 & 785 \\
\CVC & 55.6 & 790 \\
W+symst & 55.8 & 793 \\
KN40+symst+m10u+atponly & 56.0 & 796 \\
B+v\_pref+atponly & 56.2 & 798 \\
\LEO-po1 & 56.3 & 799 \\
\leanCoP & 56.3 & 800 \\
B+symst & 56.4 & 801 \\
B+symst+triv & 56.5 & 802 \\\bottomrule
\end{tabular}
\label{union0}
\end{small}
\end{table}

\section{Initial Comparison of the Advised and Original Proofs}
\label{comparison}
There are 6162 theorems  that can be proved by 
either \Vampire, \Z or \E from the original
\HOL dependencies. Their collection is denoted as $Original$. There are 5580 theorems (denoted as $Advised$) that can be proved by these ATPs
from the premises advised automatically.
It is interesting to see how these two sets of ATP proofs compare. In
this section, a basic comparison 
in terms of the number of
premises used for the ATP proofs is provided. A more involved
comparison and research of the proofs using the proof-complexity
metrics developed for \MML in~\cite{AlamaKU12} is left as an interesting
future work.
The intersection of $Original$ and $Advised$ contains 4694
theorems. Both sets of proofs are already minimized as described in
Section~\ref{proofdata}.
The proof dependencies were extracted\footnote{See 
  \url{http://mws.cs.ru.nl/~mptp/hh1/ATPdeps/deps_of_atp_proofs_from_hol_deps.txt}
  and
  \url{http://mws.cs.ru.nl/~mptp/hh1/ATPdeps/deps_from_advised_atp_proofs.txt}.}
and the number of dependencies was compared. The complete results of
this comparison are available
online,\footnote{\url{http://mws.cs.ru.nl/~mptp/hh1/ATPdeps/deps_comparison.txt}}
sorted by the difference between the length of the $Original$ proof and
the $Advised$ proof.
To make it easier to explore the differences described in the next subsections, the \Flyspeck and \HOLLight Subversion
repositories were merged into one (git) repository, and (quite
imperfectly)
HTML-ized\footnote{\url{http://mws.cs.ru.nl/~mptp/hol-flyspeck/index.html}}
by a simple heuristic Perl script. A simple CGI 
script\footnote{e.g., \url{http://mws.cs.ru.nl/~mptp/cgi-bin/browseproofs.cgi?refs=COMPLEX_MUL_CNJ}} can be used to
compare the dependencies needed for the (minimized) advised ATP proof with the
dependencies needed for the ATP proof from the original \HOLLight
premises, and also with the actual \HOLLight proof.

\subsection{Theorems Proved Only with Advice}
The list of 885 theorems proved only with advice is available 
online\footnote{\url{http://mws.cs.ru.nl/~mptp/hh1/ATPdeps/aonly_by_length.txt}}
sorted by the number of necessary premises. The last theorem in this
order
(\th{CROSS_BASIS_NONZERO})\footnote{\url{http://mws.cs.ru.nl/~mptp/cgi-bin/browseproofs.cgi?refs=CROSS_BASIS_NONZERO}}
used 34 premises for the advised ATP proof, while its \HOLLight proof
is just a single invocation of the \th{VEC3_TAC} tactic\footnote{An
interesting future work is to integrate calls to such tactics into the
learning/ATP framework, or even to learn their construction (from
similar sequences of lemmas used on similar inputs). The former task is
similar to optimizing SMT solvers and tools like
\MetiTarski.} (which however brings in 121 \HOLLight dependencies,
making re-proving difficult). The following two short examples show how the
advice can sometimes get simpler proofs.
\\
\\
1. Theorem \th{FACE_OF_POLYHEDRON_POLYHEDRON}
states that a face of a polyhedron (defined in \HOLLight generally as a finite intersection
of half-spaces) is again a polyhedron:
\begin{holnb}\holthm{
!s:real^N->bool c. polyhedron s /\ c face_of s ==> polyhedron c
}
\end{holnb}
The
\HOLLight proof\footnote{\url{http://mws.cs.ru.nl/~mptp/cgi-bin/browseproofs.cgi?refs=FACE_OF_POLYHEDRON_POLYHEDRON}}
 takes 23
lines and could not be re-played by ATPs, but a much simpler proof was found by the
AI/ATP automation, based on (a part of) the \th{FACE_OF_STILLCONVEX}
theorem: a face $t$ of any convex set
$s$ is equal to the intersection of $s$ with the affine hull of
$t$. To finish the proof, one needs just three ``obvious'' facts: Every
polyhedron is convex (\th{POLYHEDRON_IMP_CONVEX}), the intersection of two polyhedra is again a
polyhedron (\th{POLYHEDRON_INTER}), and affine hull is always a
polyhedron (\th{POLYHEDRON_AFFINE_HULL}):
\begin{holnb}\holthm{
FACE_OF_STILLCONVEX:
!s t:real^N->bool. convex s ==> 
(t face_of s <=> t SUBSET s /\ convex(s DIFF t) /\ t = (affine hull t) INTER s)
POLYHEDRON_IMP_CONVEX: !s:real^N->bool. polyhedron s ==> convex s
POLYHEDRON_INTER:
!s t:real^N->bool. polyhedron s /\ polyhedron t ==> polyhedron (s INTER t)
POLYHEDRON_AFFINE_HULL: !s. polyhedron(affine hull s)
}\end{holnb}
\vspace{1mm}
2. Theorem \th{FACE_OF_AFFINE_TRIVIAL} states that faces of affine sets
are trivial:
\begin{holnb}\holthm{
!s f:real^N->bool. affine s /\ f face_of s ==> f = \Empt{} \/ f = s
}
\end{holnb}
The \HOLLight
proof\footnote{\url{http://mws.cs.ru.nl/~mptp/cgi-bin/browseproofs.cgi?refs=FACE_OF_AFFINE_TRIVIAL}}
takes 19 lines and could not be re-played by ATPs. The advised proof
finds a simple path via previous
theorem \th{FACE_OF_DISJOINT_RELATIVE_INTERIOR} saying that nontrivial
faces are disjoint with the relative interior, and theorem
\th{RELATIVE_INTERIOR_UNIV} saying that any affine hull is equal to
its relative interior. The rest is again just use of several ``basic facts''
about the topic (skipped here):
\begin{holnb}\holthm{
FACE_OF_DISJOINT_RELATIVE_INTERIOR:
  !f s:real^N->bool. f face_of s /\ ~(f = s) ==> f INTER relative_interior s = \Empt{}

RELATIVE_INTERIOR_UNIV: !s. relative_interior(affine hull s) = affine hull s
}
\end{holnb}

\subsection{Examples of Different Proofs}
Finally, several examples are shown where the advised ATP proof differs from the
ATP proof reconstructed from the original \HOLLight dependencies. 
\\
\\
1. Theorems
\th{COMPLEX_MUL_CNJ}\footnote{\url{http://mws.cs.ru.nl/~mptp/cgi-bin/browseproofs.cgi?refs=COMPLEX_MUL_CNJ}}
and \th{COMPLEX_NORM_POW_2} 
stating the equality of squared norm to multiplication with a complex conjugate
follow easily from each other (together with the commutativity of
complex multiplication \th{COMPLEX_MUL_SYM}). 
The proof of \th{COMPLEX_MUL_CNJ} in \HOLLight (below) re-uses the longer proof of \th{COMPLEX_NORM_POW_2}.
The advised ATP proof directly uses \th{COMPLEX_NORM_POW_2}, but (likely because \th{COMPLEX_MUL_SYM} was never
used before) first unfolds the definition of complex
conjugate and then applies commutativity of
real multiplication. 

\begin{holnb}\holthm{
let COMPLEX_MUL_CNJ = prove
 (`!z. cnj z * z = Cx(norm(z)) pow 2 /\ z * cnj z = Cx(norm(z)) pow 2`,
  GEN_TAC THEN REWRITE_TAC[COMPLEX_MUL_SYM] THEN
  REWRITE_TAC[cnj; complex_mul; RE; IM; GSYM CX_POW; COMPLEX_SQNORM] THEN
  REWRITE_TAC[CX_DEF] THEN AP_TERM_TAC THEN BINOP_TAC THEN
  CONV_TAC REAL_RING);;

COMPLEX_NORM_POW_2: !z. Cx(norm z) pow 2 = z * cnj z
COMPLEX_MUL_SYM: !x y. x * y = y * x

}\end{holnb}
2. Theorem
\th{disjoint_line_interval}\footnote{\url{http://mws.cs.ru.nl/~mptp/cgi-bin/browseproofs.cgi?refs=Vol1.disjoint_line_interval}}
states that the left endpoints of two unit-long integer-ended intervals on the real line
have to coincide if the intervals share a point strictly inside them. This
suggests case analysis, which is what the longer \HOLLight
proof (omitted here) seems to do. The advisor instead gets the proof in a single stroke
by noticing a previous theorem saying that the left endpoint is
the $floor$ function which is constant for the points inside such intervals:
\begin{holnb}\holthm{
disjoint_line_interval: 
!(x:real) (y:real). integer x /\ integer y /\ 
  (? (z:real). x < z /\ z < x + &1 /\ y < z /\ z < y + &1) ==> x = y

FLOOR_UNIQUE: !x a. integer(a) /\ a <= x /\ x < a + &1 <=> (floor x = a)

}\end{holnb}
3. Theorem \th{NEGLIGIBLE_CONVEX_HULL_3}\footnote{\url{http://mws.cs.ru.nl/~mptp/cgi-bin/browseproofs.cgi?refs=NEGLIGIBLE_CONVEX_HULL_3}}
states that the convex hull of three points in $\mathbb{R}^3$ is a negligible
set. In \HOLLight this is proved from the general theorem \th{NEGLIGIBLE_CONVEX_HULL} stating this
property for any finite set of points in $\mathbb{R}^n$ with cardinality less or equal to $n$. 
Instead of justifying this precondition, a shorter proof is
found by the advised ATP that saw an analogous theorem about the
affine hull, the inclusion of the convex hull in the affine hull, and
the preservation of negligibility under inclusion.

\begin{holnb}\holthm{
let NEGLIGIBLE_CONVEX_HULL_3 = prove
 (`!a b c:real^3. negligible (convex hull {a,b,c})`,
  REPEAT GEN_TAC THEN MATCH_MP_TAC NEGLIGIBLE_CONVEX_HULL THEN
  SIMP_TAC[FINITE_INSERT; CARD_CLAUSES; FINITE_EMPTY; DIMINDEX_3] THEN
  ARITH_TAC);;

NEGLIGIBLE_CONVEX_HULL:
!s:real^N->bool. FINITE s /\ CARD(s) <= dimindex(:N) ==> negligible(convex hull s)

NEGLIGIBLE_AFFINE_HULL_3: !a b c:real^3. negligible (affine hull {a,b,c})
CONVEX_HULL_SUBSET_AFFINE_HULL: !s. (convex hull s) SUBSET (affine hull s)
NEGLIGIBLE_SUBSET:
 !s:real^N->bool t:real^N->bool. negligible s /\ t SUBSET s ==> negligible t
}\end{holnb}
4. Theorem \th{BARV_CIRCUMCENTER_EXISTS}\footnote{\url{http://mws.cs.ru.nl/~mptp/cgi-bin/browseproofs.cgi?refs=Rogers.BARV_CIRCUMCENTER_EXISTS}}
says that under certain assumptions, a particular point (circumcenter) lies in a particular set
(affine hull). The \HOLLight proof unfolds some of the assumptions and
takes 14 lines. The advisor just found a related theorem
\th{MHFTTZN3} which under the same assumptions states that the singleton
containing the circumcenter is equal to the intersection of the affine
hull with another set. The rest are two ``obvious'' facts about 
elements of intersections (\th{IN_INTER}) and  elements of singletons (\th{IN_SING}): 
\begin{holnb}\holthm{
BARV_CIRCUMCENTER_EXISTS: !V ul k. packing V /\ barV V k ul ==>
  circumcenter (set_of_list ul) IN affine hull (set_of_list ul)

 MHFTTZN3: !V ul k. packing V /\ barV V k ul ==>
   ((affine hull (voronoi_list V ul)) INTER (affine hull (set_of_list ul)) =
   { circumcenter (set_of_list ul) } )

IN_SING: !x y. x IN {y:A} <=> (x = y)
IN_INTER: !s t (x:A). x IN (s INTER t) <=> x IN s /\ x IN t
}\end{holnb}
5. An example of the reverse phenomenon (i.e., the advised proof is more
complicated than the original) is theorem
\th{BOUNDED\_CLOSURE\_EQ}\footnote{\url{http://mws.cs.ru.nl/~mptp/cgi-bin/browseproofs.cgi?refs=BOUNDED_CLOSURE_EQ}}
saying that a set in $R^n$ is bounded iff its closure is bounded.
The harder direction of the equivalence was already available as
theorem \th{BOUNDED\_CLOSURE}, and was used both by the \HOLLight and
the advised proof. The easier direction was in \HOLLight proved by
theorems \th{CLOSURE\_SUBSET} and \th{BOUNDED\_SUBSET} saying
that any set is a subset of its closure and any subset of a bounded
set is bounded. The advised proof instead went through a longer
path based on theorems \th{CLOSURE\_APPROACHABLE},
\th{IN\_BALL} and \th{CENTRE\_IN\_BALL} to show that every
element in a set is also in its closure, and then unfolded the definition of
\th{bounded} and showed that the bound on the norms of closure elements
can be used also for the original set.

\begin{holnb}\holthm{
let BOUNDED_CLOSURE_EQ = prove
 (`!s:real^N->bool. bounded(closure s) <=> bounded s`,
  GEN_TAC THEN EQ_TAC THEN REWRITE_TAC[BOUNDED_CLOSURE] THEN
  MESON_TAC[BOUNDED_SUBSET; CLOSURE_SUBSET]);;

BOUNDED_CLOSURE: !s:real^N->bool. bounded s ==> bounded(closure s)
BOUNDED_SUBSET: !s t. bounded t /\ s SUBSET t ==> bounded s
CLOSURE_SUBSET: !s. s SUBSET (closure s)

CLOSURE_APPROACHABLE: 
  !x s. x IN closure(s) <=> !e. &0 < e ==> ?y. y IN s /\ dist(y,x) < e
IN_BALL: !x y e. y IN ball(x,e) <=> dist(x,y) < e
CENTRE_IN_BALL: !x e. x IN ball(x,e) <=> &0 < e
bounded: bounded s <=> ?a. !x:real^N. x IN s ==> norm(x) <= a

}\end{holnb}

\subsection{Remarks}

The average number of 
\HOLLight proof dependencies restricted to the set of theorems re-proved by ATPs  is 
34.54, i.e., there are on average about nine times more dependencies
in a \HOLLight proof than in the corresponding ATP proof (see Table~\ref{trainingpasses}).
This perhaps casts some light on how learning-assisted ATP
currently achieves its 
performance. A large
human-constructed library like \Flyspeck is often dense/redundant
enough\footnote{As long as such libraries are human-constructed,
  they will remain imperfectly organized and redundant. No ``software
  engineering'' or other approach can prevent new shortcuts
  to be found in mathematics, unless an exhaustive (and infeasible)
  proof minimization is applied.} 
to allow short
proofs under the assumption of perfect (and thus inhuman) premise
selection. Such short proofs
can be found even by the quite exhaustive methods employed by most of
the existing ATPs. The smarter the premise selection and the stronger
the search inside the ATPs, the greater the chance that such proofs
will end up inside the ATP's time-limited search envelope. 
The outcome of using such advisors extensively could be
``better-informed'' mathematics that has shorter proofs which use
a variety of lemmas much more than the basic definitions and theorems. Whether such mathematics
is easier for human consumption is not clear.
Already now  mathematical texts sometimes
optimize proof length by lemma re-use to an extent that may make
the underlying ideas less
visible. Perhaps this is just another case where the strong automation
tools will eventually help to understand how human cognition works.

The ATP search is quite unlike the much less exhaustive search done by decision
procedures, and also unlike the human proofs, where the global economy of
dependencies is not so crucial once a fuzzy high-level path
to the goal gets some credibility. Both the human and the
decision-procedure proofs result in more redundant (``sloppier'')
proofs, which can however be more involved (complicated) than what the ATPs can achieve even
with optimal premise selection. Learning such (precise or fuzzy) high-level pathfinding 
is an interesting next challenge for large-theory AI/ATP systems. 
With the number of proofs and theory
developments to learn from available now in the \HOL/\Flyspeck,
\Mizar/\MML, and \Isabelle corpora, and the already relatively strong
performance of the ``basic'' AI/ATP methods that are presented in this
paper, these next steps seem to be worth a try.

\section{Related Work and Contributions}
\label{related}
Related work has been mentioned throughout the paper,
  and some of the papers cited provide recent overviews of various
  aspects of our work.
In particular, Blanchette's PhD thesis and~\cite{BlanchetteBPS13} give a detailed overview of
the translation methods for the (extended) \HOL logic used in
\Isabelle. See~\cite{Urban11-ate,KuhlweinLTUH12} for recent overviews of
large-theory ATP methods, and~\cite{UrbanV13} for a summary of the
work done over \MML and its AI aspects.

Automated theorem proving over large theories goes
back at least to Quaife's large
developments~\cite{Qua92-Book} with \Otter~\cite{MW97}\footnote{An interesting case is McAllester's
Ontic~\cite{McAllester89}. The whole library is
searched automatically, but the automation is fast and intentionally incomplete.} (continued to
some extent by Belinfante~\cite{Belinfante99a}).  
Most of the ATP/ITP combinations developed in 1990s used ATPs on user-restricted search space.
Examples
include the ATPs for \HOL (\textsf{Light}) by Harrison and Hurd mentioned
above, similar work for \Isabelle by
Paulson~\cite{Paulson99}, integration of CL$^{\mbox{A}}$M with \HOL~\cite{SlindGBB98} and integration of ATPs with the Omega proof assistant~\cite{Meier00}. Dahn, Wernhard and Byli{\'n}ski 
exported \Mizar/\MML into the \ILF format~\cite{Dah97}, created
(small) ATP problems from several \Mizar articles, and researched
ATP-friendly encodings of \Mizar's dependent and order-sorted type
system~\cite{Dahn98}. Large-theory ATP 
reappeared in 2002 
with Voronkov's and Riazanov's customized \Vampire answering queries over
the whole \SUMO ontology~\cite{PeaseS07}, and
Urban's \MoMM (modified \E) authoring tool~\cite{Urban06-ijait}
using all \MML lemmas for
dependently-typed subsumption of new \Mizar goals.
Since 2003, experiments with (unmodified) ATPs over large libraries
have been carried out for \MML~\cite{Urb04-MPTP0} (using machine learning for
premise selection) and for \Isabelle/\HOL (using the symbol-based \Sledgehammer
heuristic for premise selection).
A number of large-theory ATP methods and systems (e.g., \SInE,
\MaLARea, goal-oriented heuristics inside ATPs) have been developed
recently and evaluated over large-theory benchmarks and
competitions like CASC LTB and Mizar@Turing.
A comprehensive comparison of ATP and \Mizar proofs was recently done
in~\cite{AlamaKU12}.  As here, the average number of \Mizar proof
dependencies is higher than the number of ATP dependencies, however,
the difference is not as striking as for \HOLLight (a very different
method is used to get the \Mizar dependencies).

The work described here adds \HOLLight and \Flyspeck to the pool of
systems and corpora accessible to large-theory AI/ATP methods and
experiments.  A number of large-theory
techniques are re-used, sometimes the \Mizar, \Isabelle and CASC LTB
approaches are combined and adapted to the \HOLLight setting, and some
of the techniques are taken further. The theorem naming, dependency
export, problem creation, and advising required newly
implemented \HOLLight functions. The machine learning adds $k$-nearest neighbor, and the feature characterization
was improved by replacing variables in terms with their
\HOL types. A \MaLARea-like pass interleaving ATP with learning
was used to obtain as many ATP proofs as possible, and the proofs 
were postprocessed by pseudo- and cross-minimization.
Unlike in \MaLARea, this was done in a scenario
that emulates the growth of the library, i.e., no information about
the proofs of later theorems was used to train premise selection for
earlier theorems.  Motivated by the recent experiments over the
MPTP2078 benchmark, the machine learning was complemented by various
\SInE strategies used by \E and \Vampire. The strategy-scheduling
version of \E using the strategies developed for 
Mizar@Turing was tested for the first time in such large evaluation. 
A significant effort was spent to find the most orthogonal
ingredients of the final mix of premise selectors and ATPs: in total
435 different combinations were tested.
The resulting 39\% chance of
proving the next theorem without any user advice is a landmark for
a library of this size. While a similar number was achieved
in~\cite{KuhlweinLTUH12} on the much smaller MPTP2078 benchmark with a
lower time limit, only 18\% success rate was
recently reported in~\cite{AlamaKU12} for the whole \MML in this fully
push-button mode.\footnote{A similar large-scale evaluation for \Isabelle would be interesting. It is not clear whether the current ``Judgement Day'' benchmark contains goals on the same (theorem) level of granularity.} 
None of those evaluations however combined so many
methods as here. The improvement over the best method (proving
24.1\% theorems in 30s and 26.8\% in 300s) shows that such
combinations significantly improve the usability of
large-theory ATP methods for the end users.

\section{Future Work}
\label{future}
Stronger machine
learning (kernel/ensemble, etc.) methods and more suitable
characterizations (e.g., addition of model-evaluation features and
more abstract features) are
likely to further improve the performance. The prototype online
service could be made customizable by learning from users' own proofs. So far only
three ATPs are used by the service, but many other systems can 
eventually be added, possibly with various custom mappings to their logics. The
translation methods can be further experimented with: either to get a
symbol-consistent first-order translation (to allow, e.g., the
model-evaluation features), or to get less incomplete translations.
Proof reconstruction is currently work in progress. A simple and obvious
approach is to try \MESON with the minimized set of dependencies.
When it is ready, unsound translations can be added to the pool of
methods as was originally done in Isabelle \Isabelle~\cite{MengP08}.
Training ATP-internal guidance on the corpus for prototype
learning/ATP systems like \MaLeCoP will be interesting, and perhaps also further tuning of
ATP strategies for systems like \E. 

The power of the combined system probably already now makes it interesting
as a complementary semantic aid/filter for first experiments with statistical translation methods between
the informal \Flyspeck text and the \Flyspeck formalization.
The cases of machine translation (as in Google Translate) and
natural-language query answering (as in IBM Watson) have recently demonstrated 
the power of large-corpus-driven methods 
to automatically learn such translation/understanding layers from
uncurated imperfect resources such as Wikipedia. 
In other words, large bodies of
mathematics (and exact science) such as \href{http://arxiv.org}{arXiv.org} are unlikely to become 
computer-understandable by the current painstaking human encoding
efforts and additions of further and further logic complexity layers 
that increase the
formalization barrier both for humans and AI systems. Large-scale (world-knowledge-scale)
formalization for (mathematical) masses is hard to imagine as one
large ``perfectly engineered'' knowledge base in which everyone will
know perfectly well where their knowledge fits. Such attempts seem to
be as doomed as the initial attempts (in the Stone Age of Internet) to manually organize the World Wide Web in one
concise directory. 
Gradual world-scale formalization
seems more likely to happen through simpler logics that can be reasonably
crowd-sourced (e.g., as Wikipedia was), assisted by AI (learning/ATP)
methods continuously training and self-improving
on cross-linked formal/semiformal/informal corpora expressed in simple formalisms that
can be reasonably explained to such automated/AI methods.

\section{Acknowledgments}

Tom Hales helped to start the \MESON exporting work at CICM 2011, and
his interest as a leader of \Flyspeck has motivated us. Thanks to John
Harrison for discussing \MESON and related topics. Mark Adams gave us
his \HOLLight proof export data for \HOL Zero, which made it easy to
start the initial re-proving and learning experiments.  Piotr Rudnicki
has made his \Mizar workstation available for a number of experiments
(and his enthusiasm and support will be sorely missed).  Andrei
Paskevich has implemented the \Why bridge to the TFF1 format just in
time for our experiments. Finally, this work stands on the shoulders
of many ATP and ITP (in particular \HOL and \Isabelle) developers, and
tireless ATP competition organizers and standards producers.  We are
thankful to the JAR referees, PAAR 2012 referees and
also to Jasmin Blanchette for their extensive comments on the early
versions of this paper.

\bibliography{ate11}

\begin{thebibliography}{10}

\bibitem{Adams10}
Mark Adams.
\newblock Introducing {HOL Zero} - (extended abstract).
\newblock In Komei Fukuda, Joris van~der Hoeven, Michael Joswig, and Nobuki
  Takayama, editors, {\em ICMS}, volume 6327 of {\em LNCS}, pages 142--143.
  Springer, 2010.

\bibitem{AkbarpourP10}
Behzad Akbarpour and Lawrence~C. Paulson.
\newblock {MetiTarski}: An automatic theorem prover for real-valued special
  functions.
\newblock {\em J. Autom. Reasoning}, 44(3):175--205, 2010.

\bibitem{AlamaBMU11}
Jesse Alama, Kasper Brink, Lionel Mamane, and Josef Urban.
\newblock Large formal wikis: Issues and solutions.
\newblock In James~H. Davenport, William~M. Farmer, Josef Urban, and Florian
  Rabe, editors, {\em Calculemus/MKM}, volume 6824 of {\em LNCS}, pages
  133--148. Springer, 2011.

\bibitem{abs-1108-3446}
Jesse Alama, Tom Heskes, Daniel K\"{u}hlwein, Evgeni Tsivtsivadze, and Josef
  Urban.
\newblock Premise selection for mathematics by corpus analysis and kernel
  methods.
\newblock {\em J. Autom. Reasoning}, 52(2):191--213, 2014.

\bibitem{AlamaKU12}
Jesse Alama, Daniel K{\"u}hlwein, and Josef Urban.
\newblock {Automated and Human Proofs in General Mathematics: An Initial
  Comparison}.
\newblock In Bj{\o}rner and Voronkov \cite{DBLP:conf/lpar/2012}, pages 37--45.

\bibitem{DBLP:conf/cade/2008}
Alessandro Armando, Peter Baumgartner, and Gilles Dowek, editors.
\newblock {\em Automated Reasoning, 4th International Joint Conference, IJCAR
  2008, Sydney, Australia, August 12-15, 2008, Proceedings}, volume 5195 of
  {\em LNCS}. Springer, 2008.

\bibitem{Beeson01}
Michael Beeson.
\newblock Automatic derivation of the irrationality of e.
\newblock {\em J. Symb. Comput.}, 32(4):333--349, 2001.

\bibitem{Belinfante99a}
Johan G.~F. Belinfante.
\newblock On computer-assisted proofs in ordinal number theory.
\newblock {\em J. Autom. Reasoning}, 22(2):341--378, 1999.

\bibitem{BenzmullerPTF08}
Christoph Benzm{\"u}ller, Lawrence~C. Paulson, Frank Theiss, and Arnaud
  Fietzke.
\newblock {LEO-II} - a cooperative automatic theorem prover for classical
  higher-order logic (system description).
\newblock In Armando et~al. \cite{DBLP:conf/cade/2008}, pages 162--170.

\bibitem{DBLP:conf/cade/2011}
Nikolaj Bj{\o}rner and Viorica Sofronie-Stokkermans, editors.
\newblock {\em Automated Deduction - CADE-23 - 23rd International Conference on
  Automated Deduction, Wroc\l{}aw, Poland, July 31 - August 5, 2011.
  Proceedings}, volume 6803 of {\em LNCS}. Springer, 2011.

\bibitem{DBLP:conf/lpar/2012}
Nikolaj Bj{\o}rner and Andrei Voronkov, editors.
\newblock {\em Logic for Programming, Artificial Intelligence, and Reasoning -
  18th International Conference, LPAR-18, M{\'e}rida, Venezuela, March 11-15,
  2012. Proceedings}, volume 7180 of {\em LNCS}. Springer, 2012.

\bibitem{BlanchettePhd}
Jasmin~Christian Blanchette.
\newblock {\em Automatic Proofs and Refutations for Higher-Order Logic}.
\newblock PhD thesis, Fakult\"at f\"ur Informatik, Technische Universit\"at
  M\"unchen, 2012.
\newblock \url{http://www21.in.tum.de/~blanchet/phdthesis.pdf}.

\bibitem{BlanchetteBP11}
Jasmin~Christian Blanchette, Sascha B{\"o}hme, and Lawrence~C. Paulson.
\newblock Extending {Sledgehammer with SMT} solvers.
\newblock In Bj{\o}rner and Sofronie-Stokkermans \cite{DBLP:conf/cade/2011},
  pages 116--130.

\bibitem{BlanchetteBPS13}
Jasmin~Christian Blanchette, Sascha B{\"o}hme, Andrei Popescu, and Nicholas
  Smallbone.
\newblock Encoding monomorphic and polymorphic types.
\newblock In Nir Piterman and Scott~A. Smolka, editors, {\em TACAS}, volume
  7795 of {\em Lecture Notes in Computer Science}, pages 493--507. Springer,
  2013.

\bibitem{BlanchetteBN11}
Jasmin~Christian Blanchette, Lukas Bulwahn, and Tobias Nipkow.
\newblock Automatic proof and disproof in {Isabelle/HOL}.
\newblock In Cesare Tinelli and Viorica Sofronie-Stokkermans, editors, {\em
  FroCoS}, volume 6989 of {\em LNCS}, pages 12--27. Springer, 2011.

\bibitem{tff1}
Jasmin~Christian Blanchette and Andrei Paskevich.
\newblock {TFF1: The TPTP} typed first-order form with rank-1 polymorphism.
\newblock In Maria~Paola Bonacina, editor, {\em CADE}, volume 7898 of {\em
  Lecture Notes in Computer Science}, pages 414--420. Springer, 2013.

\bibitem{BlanchettePWW12}
Jasmin~Christian Blanchette, Andrei Popescu, Daniel Wand, and Christoph
  Weidenbach.
\newblock More {SPASS with Isabelle} - superposition with hard sorts and
  configurable simplification.
\newblock In Lennart Beringer and Amy~P. Felty, editors, {\em ITP}, volume 7406
  of {\em LNCS}, pages 345--360. Springer, 2012.

\bibitem{sledgehammer}
Sascha B{\"o}hme and Tobias Nipkow.
\newblock {Sledgehammer: Judgement Day}.
\newblock In J{\"u}rgen Giesl and Reiner H{\"a}hnle, editors, {\em IJCAR},
  volume 6173 of {\em LNCS}, pages 107--121. Springer, 2010.

\bibitem{Brown12}
Chad~E. Brown.
\newblock Satallax: An automatic higher-order prover.
\newblock In Gramlich et~al. \cite{DBLP:conf/cade/2012}, pages 111--117.

\bibitem{Carlson1999}
Andy Carlson, Chad Cumby, Jeff Rosen, and Dan Roth.
\newblock {The SNoW Learning Architecture}.
\newblock Technical Report UIUCDCS-R-99-2101, UIUC Computer Science Department,
  5 1999.

\bibitem{Church:SimplyTyped}
Alonzo Church.
\newblock A formulation of the simple theory of types.
\newblock {\em Journal of Symbolic Logic}, 5:56--68, 1940.

\bibitem{Dahn98}
Ingo Dahn.
\newblock Interpretation of a {M}izar-like logic in first-order logic.
\newblock In Ricardo Caferra and Gernot Salzer, editors, {\em FTP (LNCS
  Selection)}, volume 1761 of {\em LNCS}, pages 137--151. Springer, 1998.

\bibitem{Dah97}
Ingo Dahn and Christoph Wernhard.
\newblock First order proof problems extracted from an article in the {MIZAR}
  {M}athematical {L}ibrary.
\newblock In Maria~Paola Bonacina and Ulrich Furbach, editors, {\em Int.
  Workshop on First-Order Theorem Proving (FTP'97)}, RISC-Linz Report Series
  No. 97-50, pages 58--62. Johannes Kepler Universit\"at, Linz (Austria), 1997.

\bibitem{z3}
Leonardo~Mendon\c{c}a de~Moura and Nikolaj Bj{\o}rner.
\newblock {Z3: An Efficient SMT Solver}.
\newblock In C.~R. Ramakrishnan and Jakob Rehof, editors, {\em TACAS}, volume
  4963 of {\em LNCS}, pages 337--340. Springer, 2008.

\bibitem{FilliatreM07}
Jean-Christophe Filli{\^a}tre and Claude March{\'e}.
\newblock The {Why/Krakatoa/Caduceus} platform for deductive program
  verification.
\newblock In Werner Damm and Holger Hermanns, editors, {\em CAV}, volume 4590
  of {\em LNCS}, pages 173--177. Springer, 2007.

\bibitem{DBLP:conf/cade/2006}
Ulrich Furbach and Natarajan Shankar, editors.
\newblock {\em Automated Reasoning, Third International Joint Conference, IJCAR
  2006, Seattle, WA, USA, August 17-20, 2006, Proceedings}, volume 4130 of {\em
  LNCS}. Springer, 2006.

\bibitem{GelderS06}
Allen~Van Gelder and Geoff Sutcliffe.
\newblock Extending the {TPTP} language to higher-order logic with automated
  parser generation.
\newblock In Furbach and Shankar \cite{DBLP:conf/cade/2006}, pages 156--161.

\bibitem{GF92}
Michael~R. Genesereth and Richard~E. Fikes.
\newblock {Knowledge Interchange Format, Version 3.0 Reference Manual}.
\newblock Technical Report Logic-92-1, Computer Science Department, Stanford
  University, 1992.

\bibitem{DBLP:conf/cade/2012}
Bernhard Gramlich, Dale Miller, and Uli Sattler, editors.
\newblock {\em Automated Reasoning - 6th International Joint Conference, IJCAR
  2012, Manchester, UK, June 26-29, 2012. Proceedings}, volume 7364 of {\em
  LNCS}. Springer, 2012.

\bibitem{Hahnle96}
Reiner H\"ahnle, Manfred Kerber, and Christoph Weidenbach.
\newblock Common syntax of the {DFGS}chwerpunktprogramm deduction.
\newblock Technical Report TR 10/96, Fakult\"at f\"ur Informatik, Universit\"at
  Karlsruhe, Karlsruhe, Germany, 1996.

\bibitem{Hales05}
Thomas~C. Hales.
\newblock Introduction to the {F}lyspeck project.
\newblock In Thierry Coquand, Henri Lombardi, and Marie-Fran{\c{c}}oise Roy,
  editors, {\em Mathematics, Algorithms, Proofs}, number 05021 in Dagstuhl
  Seminar Proceedings, pages 1--11, Dagstuhl, Germany, 2006. Internationales
  Begegnungs- und Forschungszentrum f{\"u}r Informatik (IBFI), Schloss
  Dagstuhl, Germany.

\bibitem{Hales12}
Thomas~C. Hales.
\newblock Mathematics in the age of the {T}uring machine.
\newblock {\em Lecture Notes in Logic}, 2012.
\newblock to appear; \url{http://www.math.pitt.edu/~thales/papers/turing.pdf}.

\bibitem{HalesHMNOZ10}
Thomas~C. Hales, John Harrison, Sean McLaughlin, Tobias Nipkow, Steven Obua,
  and Roland Zumkeller.
\newblock A revision of the proof of the {Kepler Conjecture}.
\newblock {\em Discrete {\&} Computational Geometry}, 44(1):1--34, 2010.

\bibitem{Harrison96}
John Harrison.
\newblock {HOL Light}: A tutorial introduction.
\newblock In Mandayam~K. Srivas and Albert~John Camilleri, editors, {\em
  FMCAD}, volume 1166 of {\em LNCS}, pages 265--269. Springer, 1996.

\bibitem{Harrison96a}
John Harrison.
\newblock A {Mizar} mode for {HOL}.
\newblock In Joakim von Wright, Jim Grundy, and John Harrison, editors, {\em
  TPHOLs}, volume 1125 of {\em LNCS}, pages 203--220. Springer, 1996.

\bibitem{Har96}
John Harrison.
\newblock {Optimizing Proof Search in Model Elimination}.
\newblock In M.~McRobbie and J.K. Slaney, editors, {\em {Proceedings of the
  13th International Conference on Automated Deduction}}, number 1104 in LNAI,
  pages 313--327. Springer-Verlag, 1996.

\bibitem{Hindley69}
R.~Hindley.
\newblock The principal type-scheme of an object in combinatory logic.
\newblock {\em Transactions of the American Mathematical Society}, 146:29--60,
  1969.

\bibitem{HoderV11}
Krystof Hoder and Andrei Voronkov.
\newblock Sine qua non for large theory reasoning.
\newblock In Bj{\o}rner and Sofronie-Stokkermans \cite{DBLP:conf/cade/2011},
  pages 299--314.

\bibitem{Hurd99}
Joe Hurd.
\newblock {Integrating Gandalf and HOL}.
\newblock In Yves Bertot, Gilles Dowek, Andr{\'e} Hirschowitz, C.~Paulin, and
  Laurent Th{\'e}ry, editors, {\em TPHOLs}, volume 1690 of {\em LNCS}, pages
  311--322. Springer, 1999.

\bibitem{Hurd02}
Joe Hurd.
\newblock An {LCF}-style interface between {HOL} and first-order logic.
\newblock In Andrei Voronkov, editor, {\em CADE}, volume 2392 of {\em Lecture
  Notes in Computer Science}, pages 134--138. Springer, 2002.

\bibitem{hurd2003d}
Joe Hurd.
\newblock First-order proof tactics in higher-order logic theorem provers.
\newblock In Myla Archer, Ben~Di Vito, and C{\'{e}}sar Mu{\~{n}}oz, editors,
  {\em Design and Application of Strategies/Tactics in Higher Order Logics
  (STRATA 2003)}, number NASA/CP-2003-212448 in {NASA} Technical Reports, pages
  56--68, September 2003.

\bibitem{Kaliszyk07}
Cezary Kaliszyk.
\newblock Web interfaces for proof assistants.
\newblock {\em Electr. Notes Theor. Comput. Sci.}, 174(2):49--61, 2007.

\bibitem{KaliszykK13}
Cezary Kaliszyk and Alexander Krauss.
\newblock Scalable {LCF}-style proof translation.
\newblock In Sandrine Blazy, Christine Paulin-Mohring, and David Pichardie,
  editors, {\em Proc. of the 4th International Conference on Interactive
  Theorem Proving (ITP'13)}, volume 7998 of {\em LNCS}, pages 51--66. Springer
  Verlag, 2013.

\bibitem{KaliszykU12}
Cezary Kaliszyk and Josef Urban.
\newblock Initial experiments with external provers and premise selection on
  {HOL Light} corpora.
\newblock In Pascal Fontaine, Renate~A. Schmidt, and Stephan Schulz, editors,
  {\em PAAR-2012}, volume~21 of {\em EPiC Series}, pages 72--81. EasyChair,
  2013.

\bibitem{Kepler11}
Johannes Kepler.
\newblock {\em Strena seu de nive sexangula}.
\newblock Frankfurt: Gottfried. Tampach, 1611.

\bibitem{Vampire}
Laura Kov{\'a}cs and Andrei Voronkov.
\newblock First-order theorem proving and {V}ampire.
\newblock In Natasha Sharygina and Helmut Veith, editors, {\em CAV}, volume
  8044 of {\em Lecture Notes in Computer Science}, pages 1--35. Springer, 2013.

\bibitem{KuhlweinU12b}
Daniel Kuehlwein and Josef Urban.
\newblock Learning from multiple proofs: First experiments.
\newblock In Pascal Fontaine, Renate~A. Schmidt, and Stephan Schulz, editors,
  {\em PAAR-2012}, volume~21 of {\em EPiC Series}, pages 82--94. EasyChair,
  2013.

\bibitem{KuhlweinLTUH12}
Daniel K{\"u}hlwein, Twan van Laarhoven, Evgeni Tsivtsivadze, Josef Urban, and
  Tom Heskes.
\newblock Overview and evaluation of premise selection techniques for large
  theory mathematics.
\newblock In Gramlich et~al. \cite{DBLP:conf/cade/2012}, pages 378--392.

\bibitem{Loveland68}
Donald~W. Loveland.
\newblock Mechanical theorem proving by model elimination.
\newblock {\em Journal of the {ACM}}, 15(2):236--251, April 1968.

\bibitem{Loveland78}
Donald~W. Loveland.
\newblock {\em Automated Theorem Proving: A Logical Basis}.
\newblock North-Holland, Amsterdam, 1978.

\bibitem{McAllester89}
David~A. McAllester.
\newblock {\em Ontic: a knowledge representation system for mathematics}.
\newblock MIT Press, Cambridge, MA, USA, 1989.

\bibitem{McC-Prover9-URL}
William McCune.
\newblock {Prover9 and Mace4}.
\newblock \url{http://www.cs.unm.edu/~mccune/prover9/}, 2005--2010.

\bibitem{MS00}
William McCune and Olga~Shumsky Matlin.
\newblock {Ivy: A Preprocessor and Proof Checker for First-Order Logic}.
\newblock In M.~Kaufmann, P.~Manolios, and J.~Strother~Moore, editors, {\em
  {Computer-Aided Reasoning: ACL2 Case Studies}}, number~4 in Advances in
  Formal Methods, pages 265--282. Kluwer Academic Publishers, 2000.

\bibitem{MW97}
William McCune and Larry Wos.
\newblock {Otter: The CADE-13 Competition Incarnations}.
\newblock {\em Journal of Automated Reasoning}, 18(2):211--220, 1997.

\bibitem{Meier00}
Andreas Meier.
\newblock System description: Tramp: Transformation of machine-found proofs
  into nd-proofs at the assertion level.
\newblock In David~A. McAllester, editor, {\em CADE}, volume 1831 of {\em
  LNCS}, pages 460--464. Springer, 2000.

\bibitem{MengP08}
Jia Meng and Lawrence~C. Paulson.
\newblock Translating higher-order clauses to first-order clauses.
\newblock {\em J. Autom. Reasoning}, 40(1):35--60, 2008.

\bibitem{ObuaSImport}
Steven Obua and Sebastian Skalberg.
\newblock Importing hol into {Isabelle/HOL}.
\newblock In Furbach and Shankar \cite{DBLP:conf/cade/2006}, pages 298--302.

\bibitem{Paulson99}
Lawrence~C. Paulson.
\newblock A generic tableau prover and its integration with {Isabelle}.
\newblock {\em J. UCS}, 5(3):73--87, 1999.

\bibitem{sledgehammer10}
Lawrence~C. Paulson and Jasmin Blanchette.
\newblock Three years of experience with {S}ledgehammer, a practical link
  between automated and interactive theorem provers.
\newblock In {\em 8th IWIL}, 2010.
\newblock Invited talk.

\bibitem{PaulsonS07}
Lawrence~C. Paulson and Kong~Woei Susanto.
\newblock Source-level proof reconstruction for interactive theorem proving.
\newblock In Klaus Schneider and Jens Brandt, editors, {\em TPHOLs}, volume
  4732 of {\em LNCS}, pages 232--245. Springer, 2007.

\bibitem{PeaseS07}
Adam Pease and Geoff Sutcliffe.
\newblock First order reasoning on a large ontology.
\newblock In Geoff Sutcliffe, Josef Urban, and Stephan Schulz, editors, {\em
  ESARLT}, volume 257 of {\em CEUR Workshop Proceedings}. CEUR-WS.org, 2007.

\bibitem{Pitts93}
Andrew Pitts.
\newblock The {HOL} logic.
\newblock In M.~J.~C. Gordon and T.~F. Melham, editors, {\em Introduction to
  {HOL}: A theorem proving environment for higher order logic}. Cambridge
  University Press, 1993.

\bibitem{Poincare13}
Henri Poincar\'e.
\newblock {\em The foundations of science: Science and hypothesis, The value of
  science, Science and method}.
\newblock The Science Press, New York, 1913.

\bibitem{PW06}
Virgile Prevosto and Uwe Waldmann.
\newblock {{SPASS+T}}.
\newblock In G.~Sutcliffe, R.~Schmidt, and S.~Schulz, editors, {\em {ESCoR
  2006}}, volume 192 of {\em CEUR}, pages 18--33, 2006.

\bibitem{Qua92-Book}
Art Quaife.
\newblock {\em {Automated Development of Fundamental Mathematical Theories}}.
\newblock Kluwer Academic Publishers, 1992.

\bibitem{RudnickiT03}
Piotr Rudnicki and Andrzej Trybulec.
\newblock On the integrity of a repository of formalized mathematics.
\newblock In Andrea Asperti, Bruno Buchberger, and James~H. Davenport, editors,
  {\em MKM}, volume 2594 of {\em LNCS}, pages 162--174. Springer, 2003.

\bibitem{Sch02-AICOMM}
Stephan Schulz.
\newblock {E - A Brainiac Theorem Prover}.
\newblock {\em AI Commun.}, 15(2-3):111--126, 2002.

\bibitem{SlindGBB98}
Konrad Slind, Michael J.~C. Gordon, Richard~J. Boulton, and Alan Bundy.
\newblock System description: An interface between {CLAM} and {HOL}.
\newblock In Claude Kirchner and H{\'e}l{\`e}ne Kirchner, editors, {\em CADE},
  volume 1421 of {\em LNCS}, pages 134--138. Springer, 1998.

\bibitem{SB10}
Geoff Sutcliffe and Christoph Benzm{\"u}ller.
\newblock {Automated Reasoning in Higher-Order Logic using the TPTP THF
  Infrastructure}.
\newblock {\em Journal of Formalized Reasoning}, 3(1):1--27, 2010.

\bibitem{SutcliffeSCB12}
Geoff Sutcliffe, Stephan Schulz, Koen Claessen, and Peter Baumgartner.
\newblock The {TPTP} typed first-order form with arithmetic.
\newblock In Bj{\o}rner and Voronkov \cite{DBLP:conf/lpar/2012}, pages
  406--419.

\bibitem{SutcliffeSCG06}
Geoff Sutcliffe, Stephan Schulz, Koen Claessen, and Allen~Van Gelder.
\newblock Using the {TPTP} language for writing derivations and finite
  interpretations.
\newblock In Furbach and Shankar \cite{DBLP:conf/cade/2006}, pages 67--81.

\bibitem{Urban03}
Josef Urban.
\newblock Translating {M}izar for first order theorem provers.
\newblock In {\em MKM}, volume 2594 of {\em LNCS}, pages 203--215. Springer,
  2003.

\bibitem{Urb04-MPTP0}
Josef Urban.
\newblock {MPTP - Motivation, Implementation, First Experiments}.
\newblock {\em Journal of Automated Reasoning}, 33(3-4):319--339, 2004.

\bibitem{Urban06-ijait}
Josef Urban.
\newblock {M}o{M}{M} - fast interreduction and retrieval in large libraries of
  formalized mathematics.
\newblock {\em Int. J. on Artificial Intelligence Tools}, 15(1):109--130, 2006.

\bibitem{Urban06}
Josef Urban.
\newblock {MPTP} 0.2: Design, implementation, and initial experiments.
\newblock {\em J. Autom. Reasoning}, 37(1-2):21--43, 2006.

\bibitem{Urban11-ate}
Josef Urban.
\newblock {An Overview of Methods for Large-Theory Automated Theorem Proving
  (Invited Paper)}.
\newblock In Peter H\"{o}fner, Annabelle McIver, and Georg Struth, editors,
  {\em ATE Workshop}, volume 760 of {\em CEUR Workshop Proceedings}, pages
  3--8. CEUR-WS.org, 2011.

\bibitem{blistr}
Josef Urban.
\newblock {BliStr: The Blind Strategymaker}.
\newblock {\em CoRR}, abs/1301.2683, 2014.
\newblock Accepted to PAAR'14.

\bibitem{abs-1109-0616}
Josef Urban, Piotr Rudnicki, and Geoff Sutcliffe.
\newblock {ATP} and presentation service for {Mizar} formalizations.
\newblock {\em J. Autom. Reasoning}, 50:229--241, 2013.

\bibitem{UrbanS10}
Josef Urban and Geoff Sutcliffe.
\newblock Automated reasoning and presentation support for formalizing
  mathematics in {Mizar}.
\newblock In Serge Autexier, Jacques Calmet, David Delahaye, Patrick D.~F. Ion,
  Laurence Rideau, Renaud Rioboo, and Alan~P. Sexton, editors, {\em
  AISC/MKM/Calculemus}, volume 6167 of {\em LNCS}, pages 132--146. Springer,
  2010.

\bibitem{US+08}
Josef Urban, Geoff Sutcliffe, Petr Pudl{\'a}k, and Ji\v{r}\'{\i} Vysko\v{c}il.
\newblock {MaLARea SG1 - Machine Learner for Automated Reasoning with Semantic
  Guidance}.
\newblock In Armando et~al. \cite{DBLP:conf/cade/2008}, pages 441--456.

\bibitem{UrbanV13}
Josef Urban and Ji\v{r}\'{\i} Vysko\v{c}il.
\newblock Theorem proving in large formal mathematics as an emerging {AI}
  field.
\newblock In Maria~Paola Bonacina and Mark~E. Stickel, editors, {\em Automated
  Reasoning and Mathematics: Essays in Memory of William McCune}, volume 7788
  of {\em LNAI}, pages 240--257. Springer, 2013.

\bibitem{UrbanVS11}
Josef Urban, Ji\v{r}\'{\i} Vysko\v{c}il, and Petr \v{S}t\v{e}p{\'a}nek.
\newblock {MaLeCoP}: Machine learning connection prover.
\newblock In Kai Br{\"u}nnler and George Metcalfe, editors, {\em TABLEAUX},
  volume 6793 of {\em LNCS}, pages 263--277. Springer, 2011.

\bibitem{Wall2000}
Larry Wall.
\newblock {\em Programming {P}erl}.
\newblock O'Reilly \& Associates, Inc., Sebastopol, CA, USA, 3rd edition, 2000.

\bibitem{Weidenbach+99}
Christoph Weidenbach, Bijan Afshordel, Uwe Brahm, Christian Cohrs, Thorsten
  Engel, Enno Keen, Christian Theobalt, and Dalibor Topi\'{c}.
\newblock System description: {SPASS} version 1.0.0.
\newblock In {\em Automated Deduction - CADE-16}, volume 1632 of {\em LNCS},
  pages 378--382. Springer Berlin Heidelberg, 1999.

\bibitem{WeidenbachDFKSW09}
Christoph Weidenbach, Dilyana Dimova, Arnaud Fietzke, Rohit Kumar, Martin Suda,
  and Patrick Wischnewski.
\newblock {SPASS Version 3.5}.
\newblock In Renate~A. Schmidt, editor, {\em CADE}, volume 5663 of {\em LNCS},
  pages 140--145. Springer, 2009.

\bibitem{Wiedijk01}
Freek Wiedijk.
\newblock Estimating the cost of a standard library for a mathematical proof
  checker.
\newblock \url{http://www.cs.ru.nl/~freek/notes/mathstdlib2.pdf}, 2001.

\bibitem{abs-1201-3601}
Freek Wiedijk.
\newblock A synthesis of the procedural and declarative styles of interactive
  theorem proving.
\newblock {\em Logical Methods in Computer Science}, 8(1), 2012.

\end{thebibliography}
\bibliographystyle{plain}

\end{document}